\documentclass[11pt]{article}

\usepackage[final]{acl}

\usepackage{times}
\usepackage{latexsym}

\usepackage[T1]{fontenc}

\usepackage[utf8]{inputenc}

\usepackage{microtype}

\usepackage{inconsolata}

\usepackage{graphicx}
\usepackage{xspace}
\usepackage{amsfonts}
\usepackage{amsmath}

\usepackage{booktabs}
\usepackage{enumitem}
\usepackage{makecell}
\usepackage{multirow}
\usepackage{subcaption}
\usepackage{xurl}

\usepackage[table,svgnames,x11names]{xcolor}
\usepackage{soul}

\newcommand{\ours}{WhiFlash\xspace}

\title{WhiFlash: Accelerating Speculative Decoding\\ with Token-Level Cross-Paradigm Routing}

\author{
\textbf{Young D. Kwon}\quad
\textbf{Miles Williams}\quad
\textbf{Rui Li} \\
\textbf{Alexandros Kouris}\quad
\textbf{Stylianos I. Venieris} \\ \\
Samsung AI Center, Cambridge, UK \\
\small{\textbf{Correspondence:} \href{mailto:yd.kwon@samsung.com}{yd.kwon@samsung.com}}
}

\begin{document}
\maketitle
\begin{abstract}
    
The autoregressive nature of large language models (LLMs) remains a significant bottleneck for inference, particularly in complex agentic workloads. While speculative decoding (SD) accelerates inference, current approaches rely on static drafting paradigms, utilising either autoregressive drafting models for reasoning or diffusion-based parallel drafting models for structured outputs. We empirically find that drafting accuracy fluctuates dramatically within a single sequence, leaving significant performance unrealised by static paradigms and coarse-grained routing. To address this volatility, we introduce \emph{\ours}, the first cross-paradigm SD method that unifies autoregressive and diffusion-based parallel drafting under a single token-level controller. \ours adopts a fine-grained routing mechanism that employs either a lightweight entropy-based or a learned neural policy, both parametrised to provide a tunable balance between expected token gain and latency. To make high-frequency switching computationally viable, we introduce novel cache-management optimisations, \textit{Lazy Catch-up} and \textit{KV-only Prefill}, reducing switching overhead to below 7\% of per-round latency. By capitalising on the complementary strengths of fundamentally distinct drafting architectures, \ours achieves significantly higher acceptance lengths, yielding category-specific throughput gains of up to 69.6\% over the state-of-the-art autoregressive EAGLE-3 and 37.3\% over the diffusion-based DFlash.\footnote{\url{https://github.com/SamsungLabs/WhiFlash}}

\end{abstract}

\section{Introduction}
\label{sec:intro}

\begin{figure}[t]
    \centerline{\includegraphics[width=\linewidth]{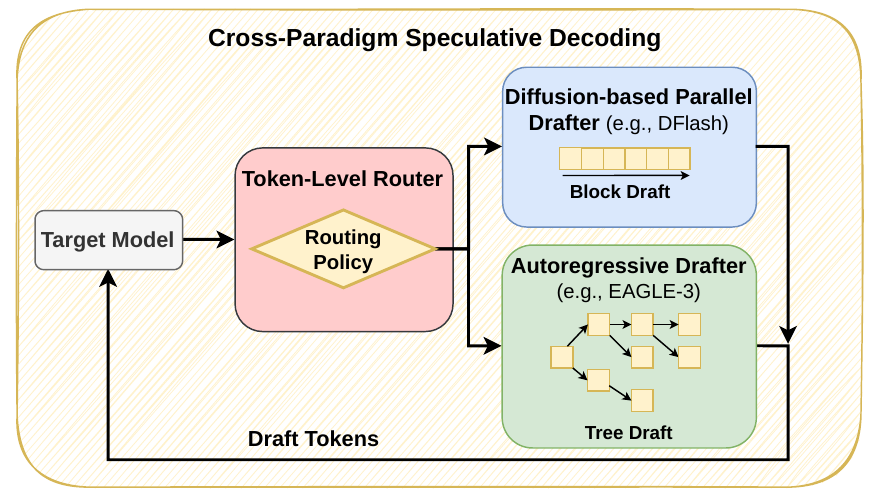}}
    \caption{\ours dynamically routes between both autoregressive and diffusion-based parallel drafters.}
    \label{fig:whiflash_diagram}
\end{figure}

Recent applications of large language models (LLMs) have evolved beyond conversational interfaces that primarily rely on parametric knowledge, towards agentic systems that interact with external environments through planning, reasoning, coding, and tool calling \citep{agarwal-etal-2025-gptoss, liu-etal-2025-deepseekv32, cao-etal-2026-qwen3codernext}. However, their efficient deployment is severely hindered by the inherent latency of autoregressive decoding. To mitigate this, speculative decoding (SD)~\citep{sd2023icml, chen-etal-2023-accelerating-large} has emerged as a primary solution. SD utilises a smaller draft model to propose tokens, which are subsequently verified in parallel by the target LLM. Nonetheless, as agentic outputs become more sophisticated, the demands placed on these draft models have diversified. Agentic workloads typically fluctuate between open-ended reasoning and highly structured phases~\citep{kim-etal-2026-cost}. Consequently, achieving peak decoding speed requires drafting mechanisms capable of dynamically handling these rapidly shifting contexts.

While a substantial body of work has advanced SD through various architectural approaches, state-of-the-art methods largely rely on static drafting paradigms. In the current landscape, standalone drafters typically excel in specific task categories while underperforming in others; autoregressive drafters~\citep{eagle3_2025neurips, wang-etal-2025-opt, liu2026talon} are highly effective in open-ended chat and novel reasoning, caching approaches~\citep{ma-etal-2025-cacheback, dumitru-etal-2025-copyspec, suffixdecoding2025neurips} excel at repetitive tasks, whereas diffusion methods~\citep{dflash2026icml,ddtree2026arxiv} dominate highly structured outputs. 

\begin{figure*}[t]
    \centerline{\includegraphics[width=0.9\textwidth]{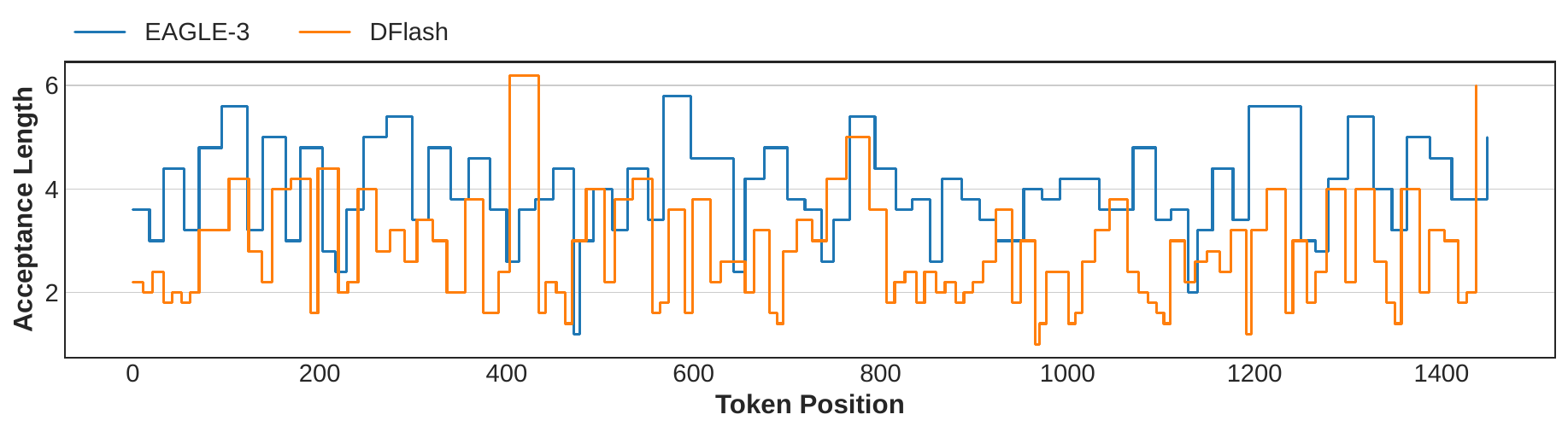}}
    \caption{Token-level acceptance lengths for EAGLE-3 and DFlash. Acceptance length fluctuates substantially throughout the response, and the positions where each drafter performs well are different.}
    \label{fig:motivation_divergence}
\end{figure*}

Crucially, beyond the cross-task performance variability, we make the key observation that drafting accuracy fluctuates aggressively even \textit{within} a single generation sequence (Figure~\ref{fig:motivation_divergence}, Section~\ref{sec:motivation}). Existing multi-drafter orchestration methods do not adequately address this token-level volatility. For example, routing mechanisms, such as HedgeSpec~\citep{hedgespec2026iclr}, formulate selection as an online learning problem designed to rapidly converge on a single, fixed expert for the remainder of the sequence. Furthermore, these frameworks operate under an assumption of architectural homogeneity, \textit{i.e.}~routing between identical types of models. Ultimately, by relying on a static drafter or coarse-grained task-level routing, current systems fail to capture token-level generation dynamics and leave significant performance gains unrealised.

In this work, we introduce \ours (Figure~\ref{fig:whiflash_diagram}), the first cross-paradigm SD method that unifies autoregressive and diffusion-based parallel drafting under a single dynamic controller. Rather than aiming to converge on a single global expert, \ours capitalises on the complementary strengths of fundamentally different drafting paradigms. To perform this with minimal latency overhead, we design a token-level router that dynamically routes each decoding step to the highest-performing draft model at run time, using either an entropy-based or neural policy. 
We make the following core contributions: 
\begin{itemize}[left=0pt]
    \item We present an analysis of intra-sequence generation dynamics, demonstrating that draft model accuracy fluctuates aggressively over token sequences and that coarse-grained, task-level routing is suboptimal. 
    \item We propose \ours, a unified speculative decoding method that dynamically orchestrates autoregressive and diffusion-based parallel drafting paradigms at maximum token-level granularity.
    \item Across a variety of tasks, \ours achieves higher acceptance lengths than static SD methods, with throughput gains of up to 69.6\% and 37.3\% over EAGLE-3 and DFlash, respectively. 
\end{itemize}

\section{Related Work}
\label{sec:related_work}

\paragraph{Standalone Draft Models.}
Recent advances in SD have primarily focused on maximising the efficacy of standalone draft models through two distinct architectural paradigms. First, \emph{autoregressive} (AR) methods capture sequential correlations between successive draft tokens and introduce dynamic, context-aware draft trees. For instance, EAGLE-3~\citep{eagle3_2025neurips} abandons top-layer feature prediction for direct token prediction via multi-layer fusion. To optimise draft token exploration, methods such as TALON~\citep{liu2026talon} employ confidence-aware token trees, while OPT-Tree~\citep{wang-etal-2025-opt} adaptively searches for a tree structure that maximises acceptance length. These AR models excel in open-ended chat and complex reasoning tasks due to their sequential-dependence modelling. Second, \emph{diffusion}-based parallel methods such as DFlash~\citep{dflash2026icml} bypass the sequential bottleneck of AR drafting by predicting tokens in blocks with a single forward pass, dominating in highly structured tasks such as coding. Despite these advances, our empirical analysis reveals that the efficacy of these individual drafters fluctuates aggressively within a single sequence (Section~\ref{sec:motivation}). Unlike these approaches that remain bound to a single paradigm, \ours is a unified method that dynamically combines parallel and AR drafting at the decoding step level.

\paragraph{Hybrid Speculative Decoding.}
To overcome the limitations of purely generative draft models, a few hybrid approaches have emerged that rely heavily on retrieval mechanisms. RASD~\citep{quan-etal-2025-rasd} fuses generative draft trees with suffix-matched retrieval trees sourced from existing datastores. SuffixDecoding~\citep{suffixdecoding2025neurips} utilises suffix trees to cache long token sequences from prompts and previous outputs, while adaptively scaling its speculation length. The paper also studies a basic hybrid variant that falls back to an autoregressive drafter when historical pattern matching yields low confidence. Furthermore, ReSpec~\citep{respec2025arxiv} transitions from heuristic retrieval to an adaptive framework via an entropy-guided trigger to decide whether to retrieve from historical context or fall back to an autoregressive drafter. While showing promise for repetitive agentic loops, retrieval inherently falters when generating novel reasoning sequences. In contrast to these retrieval-dependent systems, \ours relies either on a learned neural router or on readily available runtime metrics, rather than historical text caches, to make fine-grained paradigm-switching decisions.

\paragraph{Homogeneous Drafter Routing.}
The problem of selecting the best drafter from a pool of candidates has recently gained attention. Early approaches, such as MetaSD-UCB~\citep{kim-etal-2026-multi}, framed this as a partial-information problem, utilising multi-armed bandit algorithms to route between domain-specific AR models. To optimise compute allocation, hierarchical methods like Cascade Speculative Drafting~\citep{cascadesd2024neurips} introduce dynamic routing between models of varying sizes. More recently, HedgeSpec~\citep{hedgespec2026iclr} advanced this by framing drafter selection as a full-information online learning problem. It constructs an unbiased estimator of candidate acceptance lengths from the verified trajectory, which avoids extra target model calls. It then orchestrates a pool of domain experts using a regret-minimisation algorithm. 

However, existing routing frameworks fundamentally operate within a single architectural paradigm and optimise for a dominant expert across a task. Because the optimal generative paradigm (autoregressive versus parallel) shifts dramatically throughout a single sequence, \ours tackles a fundamentally different architectural duality. Rather than optimising to converge to a single expert, \ours introduces a token-level router that continuously toggles between architectures to capitalise on entropy variability within a sequence.

\section{Cross-Paradigm SD with \ours}
\label{sec:method}

\subsection{Motivation}
\label{sec:motivation}

Considering the two best-performing general-purpose drafting approaches, \textit{i.e.}~autoregressive and diffusion-based, we observe that both paradigms handle token-level dependencies differently. Autoregressive tree drafters generate tokens sequentially, with each position conditioning on earlier ones. In contrast, diffusion-based drafters generate a block in parallel, with positions conditioning on the prefix but not on each other.

To investigate further, we compare the speculation accuracy of autoregressive and diffusion approaches at each generation step. Figure~\ref{fig:motivation_divergence} presents an illustrative example of the acceptance lengths across an entire generated sequence for both approaches.
We observe that the acceptance length swings between one and six tokens within a single sequence, and which drafter is superior can vary substantially between positions. However, the \emph{mean} acceptance length conceals this divergence.

The volatility in acceptance length suggests that neither the autoregressive nor the diffusion drafting paradigm is superior throughout an entire sequence. This serves as the principal motivation for \ours, a new hybrid SD method that dynamically routes between heterogeneous drafters at each decoding step. By operating at this higher granularity, \ours can adaptively leverage the unique strengths of each drafting paradigm, enabling even faster decoding across a variety of tasks.

\subsection{Proposed Architecture}
\label{sec:arch}
Figure~\ref{fig:whiflash_diagram} depicts our proposed hybrid SD method, outlining the generation process. First, the \emph{target model} processes the current sequence (\textit{e.g.}~a user prompt) and samples the next token. The representations from the target model at the current decoding step are used to extract specific features. Next, the \textit{token-level router} ingests these features and uses its \textit{routing policy} to select the best-performing drafter for the upcoming speculation round. Based on this routing decision, \ours activates the corresponding speculation method. In turn, drafting proceeds through either an \textit{autoregressive drafter} which produces a tree of candidate tokens, or a \textit{diffusion-based parallel drafter} which predicts a parallel block of tokens. This sequence of candidate tokens is then returned to the target model for parallel verification. Finally, the process repeats, continuing from the last accepted token.

\subsection{Token-Level Router}
\label{sec:router_design}

To accommodate the dense intra-sequence drafting-accuracy fluctuations without inducing prohibitive computational overhead, the routing mechanism must be both highly granular and efficient. 
To this end, we introduce a \textit{token-level router}, our core module that makes dynamic decisions by means of a \textit{routing policy}. To swiftly adapt to changing contexts within a given sequence, the token-level router is designed to be invoked at every decoding step, \textit{i.e.}~once per speculation round, enabling maximum decision-making granularity. 

Given the available draft models $\{m_\text{a}, m_\text{d}\}$, where $m_\text{a}$ and $m_\text{d}$ are the autoregressive and diffusion draft models, respectively, the routing policy aims to determine which drafter to utilise at decoding step $t$, such that the acceptance length is maximised. Under this setup, we propose two types of routing policies: (1)~a runtime-informed policy that utilises available statistics for routing decisions, and (2)~a neural policy that is learned offline prior to deployment.

\subsubsection{Runtime-Informed Policy}
\label{sec:runtime_policy}

We first introduce a runtime-informed routing policy, \textbf{\ours-Entropy}, that uses the entropy of the next-token distribution from the target model as the routing signal. Our hypothesis is that entropy indicates which drafting paradigm performs better at a given step. When the next-token distribution has low entropy, the next token is more predictable, and the diffusion drafter can commit an entire block accurately whilst benefiting from its single-pass generation. When entropy is high, the next token has greater uncertainty, and the parallel drafter, which fixes a whole block at once, is more likely to mispredict. In contrast, the autoregressive drafter, which drafts sequentially and resolves one position at a time, is better suited to this regime.

Concretely, at each speculation round, the target model verifies the tokens proposed by the current drafter and produces a next-token distribution $p(x_t \mid x_{<t})$ over the vocabulary $\mathcal{V}$ for the final accepted position. The router computes the entropy, \mbox{$H_t = -\sum_{x \in \mathcal{V}} p(x \mid x_{<t}) \log p(x \mid x_{<t})$}, and selects the draft model for the next round as:
\begin{equation*}
    \pi(H_t, \tau_{\text{entropy}}) = 
    \begin{cases} 
    m_\text{a}, & \text{if } H_t > \tau_{\text{entropy}} \\ 
    m_\text{d}, & \text{otherwise} 
    \end{cases}
\end{equation*}

where $\tau_{\text{entropy}}$ is the entropy threshold. Computing the entropy requires an $O(|\mathcal{V}|)$ reduction over the target next-token probabilities at a single position, adding a softmax only when greedy decoding is used. This cost is negligible compared to a draft model forward pass, making \ours-Entropy extremely lightweight for routing during inference.

\paragraph{Calibration.} The threshold $\tau_{\text{entropy}}$ is calibrated on a small held-out set of prompts, randomly sampled and disjoint from the evaluation prompts, typically 32 to 64 prompts per category (\textit{e.g.}~Chat, Coding, Math, and Agentic), via grid search. The procedure runs \ours-Entropy on each prompt to obtain throughput and selects the $\tau_{\text{entropy}}$ that maximises mean throughput under the resulting routing decisions. Note that calibration is inexpensive, requiring no model training and completing in a few minutes on a single GPU in our experiments.

\subsubsection{Neural Policy}
\label{sec:neural_policy}

In this subsection, we introduce a learned policy, \textbf{\ours-Neural}, parametrised as a neural network $f_{\theta}$, designed around three core principles. First, our policy takes the final-layer hidden states of the target model at the current decoding step $t$, denoted by $h_t \in \mathbb{R}^d$, where $d$ is the hidden dimension of the target model. With this approach, we aim to leverage the contextual information already embedded within the target model. Second, the network directly predicts a continuous scalar, $\Delta \widehat{AL}_t = f_{\theta}(h_t)$, representing the expected difference in acceptance length between the autoregressive and the diffusion-based parallel drafter at step $t$. In contrast to a binary classification approach, this strategy enables us to quantify the expected token gain when selecting one of the drafters over the other. Finally, to ensure that the decoding step invocation frequency does not counteract drafting gains, we parameterise the router as a lightweight multilayer perceptron (MLP) regressor, consisting primarily of two linear layers.\footnote{Full architectural details are presented in Appendix \ref{sec:neural_router_architecture}.}

\paragraph{Training.} 
Given a target LLM, the neural router undergoes an offline process, comprising two phases: (1)~dataset construction and (2)~training. 

In the first phase, the training set is generated by processing a task-agnostic dataset to extract $(h_t, y_t)$ pairs at each decoding step. For the training target $y_t$, we opt to use the difference in acceptance length between the two draft models, measured in tokens. That is, $\Delta AL = AL_\text{a} - AL_\text{d}$, where $AL_\text{a}$ and $AL_\text{d}$ are acceptance lengths for autoregressive and diffusion drafting, respectively. We collect the acceptance lengths by running a complete drafting cycle per decoding step for each drafter separately, and calculate and store their differences as targets.

After the dataset construction, we proceed with the training phase, supervising the MLP on these $(h_t,  y_t)$ pairs. This objective directly optimises the network to predict the expected token gain when switching drafting methods. Overall, as our neural router is trained on this generic, task-independent dataset and then applied across various downstream tasks, its training cost becomes amortised. 

\paragraph{Hardware-Aware Routing Decision.}
During generation, we translate the router's continuous prediction into a discrete routing decision. Let $f_{\theta}$ denote the trained neural regressor. At decoding step $t$, the router predicts the expected token gain, $\Delta \widehat{AL}_t$, given the target model's hidden states $h_t$, as $\Delta \widehat{AL}_t = f_\theta(h_t)$. 

A conventional strategy would select drafting methods based on the prediction's sign, \textit{i.e.}~switch to the autoregressive drafter if \mbox{$\Delta \widehat{AL}_t > 0$}. However, this assumes both draft architectures incur identical computational costs, which is rarely true in practice. Instead, \ours relies on a hardware-aware decision function in order to determine the selected draft model, controlled through a routing threshold $\tau_{\text{neural}}$:
$$\pi(h_t, \tau_{\text{neural}}) = 
\begin{cases} m_\text{a}, & \text{if } \Delta \widehat{AL}_t > \tau_{\text{neural}} \\ 
m_\text{d}, & \text{otherwise} \end{cases}$$
The value of threshold $\tau_{\text{neural}}$ is explicitly informed by system-level metrics, such as the inference latency of each draft model and the switching latency overhead on the target hardware. 
As a result, the router only executes a paradigm switch when the predicted gain in accepted tokens reliably outweighs the asymmetric computational overhead of the alternative drafter.

\paragraph{Extension to Multiple Drafters.}

As we focus on the differing characteristics of autoregressive and parallel drafting, we formulate routing as a two-way problem. Nonetheless, the learned router can be readily adapted to more than two drafters by modelling the distribution of acceptance-length gain over $N>2$ drafters, replacing the MSE objective with a cross-entropy loss.

\subsection{System Optimisation}
\label{sec:sys_opt}
Operationalising two drafters at the token level poses two engineering questions concerning the AR drafter's KV cache: (1) when it should be updated, and (2) how the update itself should be computed. We propose lazy catch-up and KV-only prefill to address these issues, respectively.

\paragraph{Lazy Catch-up.}
The naive approach continuously keeps the KV cache for the inactive drafter up to date, performing prefill over every accepted token on every round, regardless of which drafter is chosen. Switches become instantaneous, although the inactive drafter repeatedly performs small prefill operations, losing the advantage of batched computation. Instead, we defer the updates for the inactive drafter entirely: no work is performed for the inactive drafter until a switch is committed. Upon preparing for a switch, the incoming drafter prefills its KV cache over all tokens accepted during its inactivity as a single batched operation. We refer to these tokens, accepted while a drafter was inactive and therefore missing from its KV cache, as \textit{backlog tokens}. Their quantity defines the size of the catch-up prefill at each switch.

\paragraph{KV-only Prefill.}
The computational cost of the catch-up prefill itself can be reduced further. A full forward pass on each token computes self-attention, the MLP, and the LM head; the LM head in particular is a known bottleneck for small drafters at large vocabularies~\citep{williams-etal-2026-speculative}. However, for catch-up, only the KV cache is required since the drafter does not need predictions for tokens that the target has already accepted. Therefore, we compute only the attention layer on tokens accepted by another drafter, processing only the final token in full, so that its post-MLP hidden state can feed the next draft round. With this technique, \ours keeps switching cost below 7\% of per-round latency.

\section{Experimental Setup}
\label{sec:eval}

\paragraph{Baselines.}
We employ two state-of-the-art standalone drafters, \textit{EAGLE-3}~\citep{eagle3_2025neurips} and \textit{DFlash}~\citep{dflash2026icml}. To ascertain the maximum achievable gain at different levels of routing granularity, we report three oracles. Each oracle selects the best drafter \textit{in hindsight}, using the performance of both drafters observed after generation, and is therefore an upper bound rather than a deployable method: (1) \textit{Oracle-Task} selects the single best drafter for each task/dataset; (2) \textit{Oracle-Prompt} selects the single best drafter for each prompt, representing the upper bound of every realisable prompt-level router (\textit{e.g.}~HedgeSpec~\citep{hedgespec2026iclr}) by construction; and (3) \textit{Oracle-Token} selects the best drafter at every decoding step, representing the best achievable routing quality for the two drafters.
To maintain identical sequences across drafters, Oracle-Token is computed in \texttt{float32}, while all other methods use \texttt{bfloat16}. Thus, the acceptance length is a valid token-level upper bound, but the throughput is not comparable. Accordingly, we omit the speedup.

\paragraph{Models.}
To assess the generalisation of each method, we target three open-weight LLMs from two model families. We employ \textit{Qwen3}~\citep{yang-etal-2025-qwen3} in 4B and 8B sizes, in addition to \textit{Llama-3.1-8B}~\citep{grattafiori2024llama3herdmodels}. We select these models because they are among the best-performing models in their size class and enable a direct comparison with prior work.

\paragraph{Training/Calibration.}
Our lightweight neural policy is trained offline and independently for each target model. We construct the training set following the process from Section~\ref{sec:neural_policy}. We adopt a task-agnostic corpus, specifically the Dolci Instruct SFT post-training dataset~\citep{olmo2026olmo3}; none of the 15 evaluation datasets is used for router training. Our neural router is optimised with AdamW~\citep{loshchilov-hutter-2018-decoupled} using an MSE loss with a peak learning rate of $1\times10^{-3}$. We train for a single epoch on a 25M token subset, using a batch size of 8K tokens. Because the corpus is task-independent, this cost is incurred once per target model and amortised across all downstream tasks. 
Our runtime-informed policy requires no training. We calibrate its single threshold $\tau_{\text{entropy}}$ with grid search over a held-out set of $32$ to $64$ prompts per category, randomly sampled from a held-out split disjoint from the evaluation prompts, and retain the value maximising mean throughput.

\paragraph{Evaluation.}
We evaluate across four task categories: (1) \textit{Chat}, with MT-Bench~\citep{zheng2023judging}, Alpaca~\citep{alpaca2023}, UltraChat~\citep{ding-etal-2023-enhancing}, and WildChat~\citep{wildchat2024}; (2) \textit{Math}, with GSM8K~\citep{gsm8k2021}, MATH-500~\citep{math500_2023}, AIME24, and AIME25~\citep{aime2025}; (3) \textit{Coding}, with HumanEval~\citep{humaneval2021}, MBPP~\citep{mbpp2021}, and LiveCodeBench~\citep{livecodebench2024}; and (4) \textit{Agentic}, with SWE-Bench~\citep{swebench2024}, Tau-Bench~\citep{taubench2024}, AgentBench~\citep{agentbench2023}, and AgentInstruct~\citep{agentinstruct2024}. We report full dataset statistics in Appendix~\ref{app:experimental_setup}. 
Unless otherwise stated, measurements are based on NVIDIA H100 GPUs at a batch size of 1; results on RTX 4090 GPUs are reported in Section~\ref{sec:analysis} and Appendix~\ref{app:analysis}.

\paragraph{Metrics.}
We focus our evaluation on three key metrics: (1) \textit{mean acceptance length} (AL), the average number of tokens accepted per target verification round; (2) \emph{throughput} (TPS), the number of generated tokens per second in wall-clock time; and (3) \emph{speedup}, the throughput ratio versus plain autoregressive decoding for the same target model. All measurements are based on the decoding stage.

\section{Results}\label{sec:results}

\begin{table*}[t]
  \addtolength{\tabcolsep}{-1pt}
  \scriptsize
  \centering
  \begin{tabular}{ll ccccc ccccc ccccc}
    \toprule
     & & \multicolumn{2}{c}{\textbf{Chat}} & \multicolumn{2}{c}{\textbf{Math}} & \multicolumn{2}{c}{\textbf{Coding}} & \multicolumn{2}{c}{\textbf{Agentic}} & \multicolumn{2}{c}{\textbf{Total Avg.}} \\
    \cmidrule(lr){3-4}\cmidrule(lr){5-6}\cmidrule(lr){7-8}\cmidrule(lr){9-10}\cmidrule(lr){11-12}
    \textbf{Model} & \textbf{Method} & AL & Speedup ($\times$) & AL & Speedup ($\times$) & AL & Speedup ($\times$) & AL & Speedup ($\times$) & AL & Speedup ($\times$) \\
    \midrule
    \multirow{7}{*}{Qwen3-4B} & EAGLE-3 & 5.21 & 2.58 & 5.44 & 2.94 & 5.25 & 2.79 & 3.99 & 2.02 & 4.95 & 2.57 \\
    & DFlash           & 3.43 & 2.08 & 7.22 & 4.75 & 6.45 & 4.29 & 3.56 & 2.34 & 5.08 & 3.30 \\
    \cmidrule{2-12}
    & Oracle-Task      & 5.21 & 2.58 & 7.22 & 4.75 & 6.45 & 4.29 & 3.99 & 2.34 & 5.67 & 3.44 \\
    & Oracle-Prompt   & 5.29 & 2.76 & 7.22 & 5.00 & 6.49 & 4.45 & 4.01 & 2.41 & 5.71 & 3.60 \\
    & Oracle-Token & 5.68 & - & 8.36 & - & 7.79 & - & 4.81 & - & 6.59 & - \\
    \cmidrule{2-12}
    & WhiFlash-Entropy & 5.19 & 2.69 & 7.36 & 4.90 & 6.73 & 4.46 & 4.22 & 2.41 & 5.82 & 3.56 \\
    & WhiFlash-Neural  & \textbf{5.41} & \textbf{2.78} & \textbf{7.66} & \textbf{4.98} & \textbf{6.97} & \textbf{4.57} & \textbf{4.40} & \textbf{2.42} & \textbf{6.05} & \textbf{3.63} \\
    \midrule
    \multirow{7}{*}{Qwen3-8B}  & EAGLE-3          & 5.24 & 2.73 & 5.58 & 3.13 & 5.29 & 2.90 & 4.42 & 2.38 & 5.12 & 2.78 \\
 & DFlash           & 3.40 & 2.12 & 7.38 & 5.23 & 6.56 & 4.44 & 4.12 & 2.62 & 5.29 & 3.55 \\
 \cmidrule{2-12}
 & Oracle-Task      & 5.24 & 2.73 & 7.38 & 5.23 & 6.56 & 4.44 & 4.45 & 2.62 & 5.86 & 3.71 \\
 & Oracle-Prompt   & 5.34 & 2.95 & 7.39 & 5.20 & 6.63 & 4.63 & 4.53 & 2.72 & 5.93 & 3.82 \\
 & Oracle-Token & 5.72 & - & 8.56 & - & 7.98 & - & 5.26 & - & 6.81 & - \\
 \cmidrule{2-12}
 & WhiFlash-Entropy & 5.24 & 2.84 & 7.54 & 5.28 & 6.97 & 4.64 & 4.78 & 2.73 & 6.08 & 3.82 \\
 & WhiFlash-Neural  & \textbf{5.47} & \textbf{2.92} & \textbf{7.72} & \textbf{5.30} & \textbf{7.20} & \textbf{4.67} & \textbf{4.90} & \textbf{2.78} & \textbf{6.26} & \textbf{3.87} \\
    \midrule
    \multirow{7}{*}{Llama3.1-8B}  & EAGLE-3          & 4.99 & 2.22 & 5.25 & 2.42 & 5.62 & 2.71 & 4.28 & 1.95 & 4.99 & 2.30 \\
 & DFlash           & 3.62 & 2.30 & 4.49 & 2.88 & 4.65 & 3.04 & 3.28 & 2.10 & 3.97 & 2.55 \\
 \cmidrule{2-12}
 & Oracle-Task      & 4.99 & 2.30 & 5.25 & 2.88 & 5.62 & 3.04 & 4.28 & 2.16 & 4.99 & 2.57 \\
 & Oracle-Prompt    & 4.44 & 2.42 & 4.85 & 2.95 & 4.92 & 3.08 & 3.93 & 2.23 & 4.51 & 2.64 \\
 & Oracle-Token & 5.30 & - & 5.95 & - & 6.76 & - & 4.58 & - & 5.36 & - \\
 \cmidrule{2-12}
 & WhiFlash-Entropy & 4.97 & 2.34 & 5.34 & 2.89 & 5.58 & \textbf{3.09} & 4.30 & \textbf{2.17} & 5.01 & \textbf{2.59} \\
 & WhiFlash-Neural  & \textbf{5.02} & \textbf{2.37} & \textbf{5.48} & \textbf{2.88} & \textbf{5.79} & 3.07 & \textbf{4.45} & 2.16 & \textbf{5.14} & \textbf{2.59} \\
    \bottomrule
  \end{tabular}
  \caption{Per-category acceptance length (AL) and speedup over vanilla AR decoding, for three target models across four task categories. Oracle rows are hindsight upper bounds. \textbf{Bold} denotes the best deployable method. Notably, the Oracle-Token speedup is omitted as it requires \texttt{float32} inference, preventing comparison with \texttt{bfloat16} inference.
  }
  \label{tab:throughput_main}
\end{table*}

\paragraph{WhiFlash consistently outperforms both DFlash and EAGLE-3 standalone drafters.}
Table~\ref{tab:throughput_main} reports mean AL per category for the standalone drafters, the three oracles, and both \ours routers. 
Neither standalone drafter is uniformly best. By mean AL, EAGLE-3 leads on Chat and Agentic across all models and DFlash leads on Math and Coding for Qwen models, whereas by speedup, DFlash leads Math, Coding, and Agentic on all models, since its lower per-round latency turns even a lower AL into a higher TPS.
This highlights that a router must intelligently select the appropriate drafter; otherwise, it risks harming decoding performance.
Oracle-Token, the ideal upper bound of token-level routing, sits above both standalone drafters and the coarser oracles. This quantifies the AL that only token-level routing reaches and that coarser routing leaves unrealised.

\ours narrows this gap substantially: averaged across categories, \ours-Neural attains a mean AL of 6.26 tokens on Qwen3-8B, a 22.3\% gain over EAGLE-3 and an 18.5\% gain over DFlash. It also surpasses Oracle-Task by 6.8\% and Oracle-Prompt by 5.7\% and closes much of the gap (short of merely 8.0\%) to the performance ceiling represented by Oracle-Token. Notably, these gains are obtained on tasks the neural router never sees during training: \ours-Neural is trained solely on a generic post-training corpus, so all 15 evaluation datasets lie outside its training distribution.\footnote{Training completes in under one hour per target model on a single H100 (see Table~\ref{tab:app:router_training_cost}, Appendix~\ref{app:experimental_setup}).}

\paragraph{\ours throughput surpasses optimal task- and prompt-level routing strategies.}
Table~\ref{tab:throughput_main} reports decoding throughput per category for each target model, with the AL advantage carrying through to wall-clock gains. We report gains relative to DFlash as it is the stronger standalone baseline at this scale. 
For Qwen3-4B, \ours-Entropy improves average throughput over DFlash by $7.8\%$ and \ours-Neural by $10.0\%$, while the corresponding gains for Qwen3-8B are $7.7\%$ and $9.1\%$. Both routers are calibrated per model and per category. The learned policy proves to be the stronger of the two, as it conditions on hidden states from the target model, capturing routing signals that a calibrated entropy threshold cannot.
The gains are largest where the standalone drafters are weakest: relative to DFlash, \ours improves throughput on Chat by up to $37.3\%$ (Qwen3-8B), the category where diffusion drafting is least effective. Compared to EAGLE-3, it reaches up to 69.6\% on Math, 63.4\% on Coding, and 19.9\% on Agentic tasks (Qwen3-4B).

Crucially, on both Qwen3 target models, \ours-Neural outperforms Oracle-Prompt, which represents the hindsight upper bound of any prompt-level router, including HedgeSpec~\citep{hedgespec2026iclr}.
These performance gains reveal that the token-level ceiling lies above what is achievable by purely prompt-level routing strategies. This finding supports our hypothesis that the optimal routing paradigm operates on a token-level basis, rather than at the prompt-level (Section~\ref{sec:motivation}).

\section{Analysis}
\label{sec:analysis}

\begin{table}[t]
  \small
  \centering
  \begin{tabular}{p{1.1cm} p{1.1cm}p{1.1cm}p{1.1cm}p{1.1cm}}
    \toprule
    $\boldsymbol{\tau}_{\textbf{neural}}$ & \textbf{Chat} & \textbf{Math} & \textbf{Coding} & \textbf{Agentic} \\
    \midrule
    $-1.0$           & 149.5 & 238.4 & 223.1 & 136.7 \\
    $-0.5$           & 147.2 & 247.7 & 226.7 & 137.4 \\
    $\phantom{-}0.0$ & 146.1 & 249.1 & 223.7 & 140.0 \\
    $\phantom{-}\textbf{0.5}$ & 148.1 & 250.6 & 228.5 & \textbf{142.7} \\
    $\phantom{-}\textbf{1.0}$ & \textbf{150.0} & \textbf{263.4} & \textbf{237.5} & 140.6 \\
    $\phantom{-}1.5$ & 148.4 & 262.4 & 235.8 & 139.4 \\
    $\phantom{-}2.0$ & 143.4 & 256.2 & 221.9 & 134.2 \\
    \bottomrule
  \end{tabular}
  \caption{Threshold sensitivity analysis with respect to the throughput of Qwen3-8B with \ours-Neural.}
  \label{tab:threshold_sensitivity}
\end{table}

\paragraph{\ours is robust across a wide threshold range.}
Table~\ref{tab:threshold_sensitivity} shows per-category throughput of
\ours-Neural on Qwen3-8B as the routing threshold $\tau_{\text{neural}}$ varies.
For \ours-Neural, the best throughput is observed within a tight band over $\tau_{\text{neural}} \in [0.5, 1.5]$. With Qwen3-8B, mean throughput stays above $184.5$ tokens/s, within 7\% of the best setting while outperforming DFlash throughout. It degrades only as the threshold approaches the extremes that recover the standalone behaviour.
Per-category calibration, which selects $\tau_{\text{neural}}$ separately per-category, improves over the best single global threshold. For Qwen3-4B, it reaches $193.9$~tokens/s, yielding a $10.0\%$ gain over DFlash and $41.1\%$ over EAGLE-3.

\begin{table}[t]
\small
  \centering
  \begin{tabular}{lccc}
    \toprule
    \textbf{Model} & \textbf{Lazy catch-up} & \textbf{TPS} & \textbf{\% Increase} \\
    \midrule
    \multirow{2}{*}{Qwen3-4B} & No & 168.3 & \phantom{0}-- \\
    & Yes & 190.9 & 13.4\% \\
    \midrule
    \multirow{2}{*}{Qwen3-8B} & No & 172.7 & \phantom{0}-- \\
    & Yes & 193.7 & 12.1\% \\
    \midrule
    \multirow{2}{*}{Llama3.1-8B} & No & 161.5 & \phantom{0}-- \\
    & Yes & 183.3 & 13.5\% \\
    \bottomrule
  \end{tabular}

  \caption{
  Lazy catch-up improves throughput without altering token acceptance through deferring KV updates.
  }
  \label{tab:lazy_catchup}
\end{table}

\paragraph{Lazy catch-up improves throughput at no accuracy cost.}
Table~\ref{tab:lazy_catchup} isolates the effect of the lazy catch-up scheme (Section~\ref{sec:sys_opt}). The naive alternative continuously updates the KV cache of the inactive drafter. This incurs a per-token prefill cost on every round, regardless of the routing decision, ultimately losing the benefit of batched computation. In contrast, lazy catch-up, which defers these updates until the router commits to a switch, performs only a single batched catch-up prefill. This avoids the redundant per-token cost and yields a throughput improvement of 12.1--13.5\% on average. The mean AL and downstream task accuracy remain unaffected, since the tokens accepted by the target are identical under both schemes.

\paragraph{KV-only prefill keeps switching cost below 7\% of per-round latency.}
We profile the cost of a cross-paradigm switch, which comprises the prefill of the backlog tokens for the EAGLE-3 drafter. The KV-only prefill of Section~\ref{sec:sys_opt} computes attention over the backlog tokens and continues processing only the final token in full. This avoids invoking the transformer MLP layer and LM head for tokens that have already been accepted by the target model, reducing the switching cost by $15\%$.

We then benchmark the switching cost by running all prompts across all tasks. We observe that the average number of backlog tokens is around 80--90 across different tasks, and that switching overhead grows linearly with the number of backlog tokens. However, per-round latency also increases, keeping the relative cost bounded.
Table~\ref{tab:switching_cost} presents the switching cost considering different sequence lengths for both the base KV cache and backlog. Notably, this remains below $7\%$ of the per-round latency across all instances, and around $3$--$5\%$ at the representative 80--90 token backlog. This demonstrates that operating two drafters does not erode the throughput gains yielded by the higher AL.

\begin{table}[t]
\small
  \centering
  \begin{tabular}{l rrrrr}
    \toprule
     & \multicolumn{5}{c}{\textbf{Backlog Tokens}}
    \\
    \cmidrule(lr){2-6}
    \textbf{Prefill Tokens} & 1 & 64 & 128 & 512 & 1024 \\
    \midrule
    512 & 3.3\% & 3.8\% & 3.8\% & 3.8\% & 4.9\% \\
    1024 & 3.3\% & 3.8\% & 3.8\% & 3.9\% & 5.2\% \\
    2048 & 3.2\% & 3.8\% & 3.8\% & 4.1\% & 5.6\% \\
    4096 & 2.8\% & 3.7\% & 3.7\% & 4.2\% & \textbf{5.8\%} \\
    8192 & 1.9\% & 3.4\% & 3.4\% & 3.8\% & 5.7\% \\
    \bottomrule
  \end{tabular}
  \caption{Profiling results for switching cost (\% w.r.t. per-round latency) with different base KV lengths and backlog sequence lengths for Qwen3-8B. 
  }
  \label{tab:switching_cost}
\end{table}

\paragraph{The neural router adds negligible decoding step overhead.}
We now investigate the overhead of operationalising our lightweight neural router. Note that \ours-Entropy relies on the runtime-informed policy, whose routing cost is the single-position softmax followed by entropy computation. Therefore, the overhead is negligible, taking 0.10 ms (0.24\% of the per-round latency). 
\ours-Neural adds a single MLP execution per decoding step. Table~\ref{tab:router_latency} shows that this execution consistently accounts for less than 1\% of the per-round latency, below the switching cost characterised above and comfortably within the throughput gains of Table~\ref{tab:throughput_main}. The computational cost of the two linear layers in the neural router is negligible relative to a draft forward pass. Consequently, the learned policy improves routing quality over \ours-Entropy at no meaningful runtime penalty.

\begin{table}[t]
\small
\centering
\begin{tabular}{lrrr}
\toprule
\textbf{Model} & \textbf{Parameters} & \textbf{Latency} & \textbf{\% Drafting} \\
\midrule
Qwen3-4B    & 0.82M & 0.29 ms & 0.73\% \\
Qwen3-8B    & 2.10M & 0.30 ms & 0.78\% \\
Llama3.1-8B & 2.10M & 0.30 ms & 0.89\% \\
\bottomrule
\end{tabular}
\caption{Neural router characteristics, including the average latency relative to the overall drafting round.}
\label{tab:router_latency}
\end{table}

\begin{table}[t]
  \addtolength{\tabcolsep}{-1.5pt}
  \small
  \centering
  \begin{tabular}{l cc}
    \toprule
    \textbf{Method} & \textbf{Memory (GB)} & \textbf{\% Increase} \\
    \midrule
    Qwen3-8B (No SD) & 15.26 & \phantom{000}-- \\
    EAGLE-3           & 16.39 & \phantom{0}7.4\% \\
    DFlash            & 17.62 & 15.4\% \\
    \ours-Entropy/Neural             & 18.41 & 20.6\% \\
    \bottomrule
  \end{tabular}
  \caption{Peak memory footprint of EAGLE-3, DFlash, and \ours. The memory overhead from \ours over a diffusion-only system (DFlash) is $0.79$ GB, merely $5.2\%$ of the target model memory footprint.}
  \label{tab:memory}
\end{table}

\paragraph{\ours adds single-digit memory overhead over a diffusion drafter.}
\ours holds two drafters and the neural router in memory, incurring a greater memory footprint than a single-drafter regime. To understand the extent of this, we examine the peak memory footprint of EAGLE-3, DFlash, and \ours. Table~\ref{tab:memory} reports the peak memory usage for each approach with Qwen3-8B and $2048$ maximum new tokens. Overall, \ours incurs merely 5.2\% memory overhead compared to DFlash. The overhead is minimal since the router is a two-layer MLP and the autoregressive drafter is a single transformer layer. In comparison to the target model and diffusion drafter, both are relatively lightweight. Therefore, \ours offers both drafting paradigms at a memory cost close to that of a single diffusion drafter.

\paragraph{The gains of \ours are robust across hardware.}
We repeat our evaluation on an NVIDIA RTX 4090 (see Table~\ref{tab:rtx4090_speedup} in Appendix~\ref{app:analysis}). The gains closely track those on the H100: \ours-Neural improves throughput over DFlash by up to $38.1\%$ and over EAGLE-3 by up to $69.8\%$ on the RTX 4090, against $37.3\%$ and $69.6\%$ on the H100. Overall, \ours remains the strongest deployable method.

\section{Conclusion}
\label{sec:conclusion}
We proposed \ours, a speculative decoding method that routes between autoregressive and diffusion-based parallel drafters at each decoding step, driven by the observation that no single paradigm dominates an entire sequence. \ours leverages either a virtually cost-free runtime-informed policy (\ours-Entropy) or a lightweight neural policy (\ours-Neural), and further system optimisations (lazy catch-up with KV-only prefill) to keep switching cost below 7\% of per-round latency. \ours improves throughput by up to 37.3\% over DFlash on Chat and by up to 69.6\% over EAGLE-3 on Math, achieving a high throughput that task-level and prompt-level routing cannot reach.

\section*{Limitations}
\ours maintains two drafters concurrently with a lightweight router, increasing memory overhead relative to a single-drafter system. However, we highlight that the autoregressive drafter is a single layer, so the added memory over a diffusion-only baseline is marginal.
This is also mitigated by our lazy catch-up and KV-only prefill, which keep switching cost below 7\% of per-round latency.
We note our evaluation focuses on the 4B to 8B scale, as no larger dense target model currently has both an EAGLE-3 and a DFlash checkpoint publicly available. Since no component of \ours is scale-specific and our gains are consistent across both scales, we expect them to carry over as such drafter pairs become available.
Finally, \ours routes each sequence independently; it fits naturally into continuous-batching engines such as vLLM and SGLang, which we leave as an extension of this work.

\bibliography{custom, anthology-1, anthology-2}

\begin{thebibliography}{40}
\providecommand{\natexlab}[1]{#1}

\bibitem[{Agarwal et~al.(2025)Agarwal, Ahmad, Ai, Altman, Applebaum, Arbus, Arora, Bai, Baker, Bao, Barak, Bennett, Bertao, Brett, Brevdo, Brockman, Bubeck, Chang, Chen, Chen, Cheung, Clark, Cook, Dukhan, Dvorak, Fives, Fomenko, Garipov, Georgiev, Glaese, Gogineni, Goucher, Gross, Guzman, Hallman, Hehir, Heidecke, Helyar, Hu, Huet, Huh, Jain, Johnson, Koch, Kofman, Kundel, Kwon, Kyrylov, Le, Leclerc, Lennon, Lessans, Lezcano-Casado, Li, Li, Lin, Liss, Lily, Liu, Liu, Lu, Lu, Martinovic, McCallum, McGrath, McKinney, McLaughlin, Mei, Mostovoy, Mu, Myles, Neitz, Nichol, Pachocki, Paino, Palmie, Pantuliano, Parascandolo, Park, Pathak, Paz, Peran, Pimenov, Pokrass, Proehl, Qiu, Raila, Raso, Ren, Richardson, Robinson, Rotsted, Salman, Sanjeev, Schwarzer, Sculley, Sikchi, Simon, Singhal, Song, Stuckey, Sun, Tillet, Toizer, Tsimpourlas, Vyas, Wallace, Wang, Wang, Watkins, Weil, Wendling, Whinnery, Whitney, Wong, Yang, Yang, Yasunaga, Ying, Zaremba, Zhan, Zhang, Zhang, Zhang, and Zhao}]{agarwal-etal-2025-gptoss}
Sandhini Agarwal, Lama Ahmad, Jason Ai, Sam Altman, Andy Applebaum, Edwin Arbus, Rahul~K. Arora, Yu~Bai, Bowen Baker, Haiming Bao, Boaz Barak, Ally Bennett, Tyler Bertao, Nivedita Brett, Eugene Brevdo, Greg Brockman, Sebastien Bubeck, Che Chang, Kai Chen, and 106 others. 2025.
\newblock \href {https://arxiv.org/abs/2508.10925} {{gpt-oss-120b \& gpt-oss-20b Model Card}}.
\newblock \emph{Preprint}, arXiv:2508.10925.

\bibitem[{{AIME}(2025)}]{aime2025}
{AIME}. 2025.
\newblock {AIME} problems and solutions.
\newblock \url{https://artofproblemsolving.com/wiki/index.php/AIME Problems and Solutions}.

\bibitem[{Austin et~al.(2021)Austin, Odena, Nye, Bosma, Michalewski, Dohan, Jiang, Cai, Terry, Le, and Sutton}]{mbpp2021}
Jacob Austin, Augustus Odena, Maxwell Nye, Maarten Bosma, Henryk Michalewski, David Dohan, Ellen Jiang, Carrie Cai, Michael Terry, Quoc Le, and Charles Sutton. 2021.
\newblock \href {https://arxiv.org/abs/2108.07732} {{Program Synthesis with Large Language Models}}.
\newblock \emph{Preprint}, arXiv:2108.07732.

\bibitem[{Cao et~al.(2026)Cao, Chen, Chen, Cui, Feng, Hui, Jing, Li, Li, Lin, Ma, Shum, Wang, Wei, Yang, Zhang, Zhang, Zhang, Zhao, and Zhou}]{cao-etal-2026-qwen3codernext}
Ruisheng Cao, Mouxiang Chen, Jiawei Chen, Zeyu Cui, Yunlong Feng, Binyuan Hui, Yuheng Jing, Kaixin Li, Mingze Li, Junyang Lin, Zeyao Ma, Kashun Shum, Xuwu Wang, Jinxi Wei, Jiaxi Yang, Jiajun Zhang, Lei Zhang, Zongmeng Zhang, Wenting Zhao, and Fan Zhou. 2026.
\newblock \href {https://arxiv.org/abs/2603.00729} {{Qwen3-Coder-Next Technical Report}}.
\newblock \emph{Preprint}, arXiv:2603.00729.

\bibitem[{Chen et~al.(2023)Chen, Borgeaud, Irving, Lespiau, Sifre, and Jumper}]{chen-etal-2023-accelerating-large}
Charlie Chen, Sebastian Borgeaud, Geoffrey Irving, Jean-Baptiste Lespiau, Laurent Sifre, and John Jumper. 2023.
\newblock \href {https://arxiv.org/abs/2302.01318} {{Accelerating Large Language Model Decoding with Speculative Sampling}}.
\newblock \emph{Preprint}, arXiv:2302.01318.

\bibitem[{Chen et~al.(2026)Chen, Liang, and Liu}]{dflash2026icml}
Jian Chen, Yesheng Liang, and Zhijian Liu. 2026.
\newblock \href {https://openreview.net/forum?id=Oz335dV48X} {{DF}lash: Block diffusion for flash speculative decoding}.
\newblock In \emph{Forty-third International Conference on Machine Learning}.

\bibitem[{Chen et~al.(2021)Chen, Tworek, Jun, Yuan, de~Oliveira~Pinto, Kaplan, Edwards, Burda, Joseph, Brockman, Ray, Puri, Krueger, Petrov, Khlaaf, Sastry, Mishkin, Chan, Gray, Ryder, Pavlov, Power, Kaiser, Bavarian, Winter, Tillet, Such, Cummings, Plappert, Chantzis, Barnes, Herbert-Voss, Guss, Nichol, Paino, Tezak, Tang, Babuschkin, Balaji, Jain, Saunders, Hesse, Carr, Leike, Achiam, Misra, Morikawa, Radford, Knight, Brundage, Murati, Mayer, Welinder, McGrew, Amodei, McCandlish, Sutskever, and Zaremba}]{humaneval2021}
Mark Chen, Jerry Tworek, Heewoo Jun, Qiming Yuan, Henrique~Ponde de~Oliveira~Pinto, Jared Kaplan, Harri Edwards, Yuri Burda, Nicholas Joseph, Greg Brockman, Alex Ray, Raul Puri, Gretchen Krueger, Michael Petrov, Heidy Khlaaf, Girish Sastry, Pamela Mishkin, Brooke Chan, Scott Gray, and 39 others. 2021.
\newblock \href {https://arxiv.org/abs/2107.03374} {{Evaluating Large Language Models Trained on Code}}.
\newblock \emph{Preprint}, arXiv:2107.03374.

\bibitem[{Chen et~al.(2024)Chen, Yang, Lin, Sun, Chang, and Huang}]{cascadesd2024neurips}
Ziyi Chen, Xiaocong Yang, Jiacheng Lin, Chenkai Sun, Kevin Chen-Chuan Chang, and Jie Huang. 2024.
\newblock \href {https://doi.org/10.52202/079017-2738} {{Cascade Speculative Drafting for Even Faster {LLM} Inference}}.
\newblock In \emph{Advances in Neural Information Processing Systems}, volume~37, pages 86226--86242. Curran Associates, Inc.

\bibitem[{Cobbe et~al.(2021)Cobbe, Kosaraju, Bavarian, Chen, Jun, Kaiser, Plappert, Tworek, Hilton, Nakano, Hesse, and Schulman}]{gsm8k2021}
Karl Cobbe, Vineet Kosaraju, Mohammad Bavarian, Mark Chen, Heewoo Jun, Lukasz Kaiser, Matthias Plappert, Jerry Tworek, Jacob Hilton, Reiichiro Nakano, Christopher Hesse, and John Schulman. 2021.
\newblock \href {https://arxiv.org/abs/2110.14168} {{Training Verifiers to Solve Math Word Problems}}.
\newblock \emph{Preprint}, arXiv:2110.14168.

\bibitem[{Ding et~al.(2023)Ding, Chen, Xu, Qin, Hu, Liu, Sun, and Zhou}]{ding-etal-2023-enhancing}
Ning Ding, Yulin Chen, Bokai Xu, Yujia Qin, Shengding Hu, Zhiyuan Liu, Maosong Sun, and Bowen Zhou. 2023.
\newblock \href {https://doi.org/10.18653/v1/2023.emnlp-main.183} {Enhancing chat language models by scaling high-quality instructional conversations}.
\newblock In \emph{Proceedings of the 2023 Conference on Empirical Methods in Natural Language Processing}, pages 3029--3051, Singapore. Association for Computational Linguistics.

\bibitem[{Dumitru et~al.(2025)Dumitru, Yang, Yadav, and Surdeanu}]{dumitru-etal-2025-copyspec}
Razvan-Gabriel Dumitru, Minglai Yang, Vikas Yadav, and Mihai Surdeanu. 2025.
\newblock \href {https://doi.org/10.18653/v1/2025.emnlp-main.1337} {{C}opy{S}pec: Accelerating {LLM}s with speculative copy-and-paste}.
\newblock In \emph{Proceedings of the 2025 Conference on Empirical Methods in Natural Language Processing}, pages 26301--26332, Suzhou, China. Association for Computational Linguistics.

\bibitem[{Ettinger et~al.(2025)Ettinger, Bertsch, Kuehl, Graham, Heineman, Groeneveld, Brahman, Timbers, Ivison, Morrison, Poznanski, Lo, Soldaini, Jordan, Chen, Noukhovitch, Lambert, Walsh, Dasigi, Berry, Malik, Shah, Geng, Arora, Gupta, Anderson, Xiao, Murray, Romero, Graf, Asai, Bhagia, Wettig, Liu, Rangapur, Anastasiades, Huang, Schwenk, Trivedi, Magnusson, Lochner, Liu, Miranda, Sap, Morgan, Schmitz, Guerquin, Wilson, Huff, Bras, Xin, Shao, Skjonsberg, Shen, Li, Wilde, Pyatkin, Merrill, Chang, Gu, Zeng, Sabharwal, Zettlemoyer, Koh, Farhadi, Smith, and Hajishirzi}]{olmo2026olmo3}
Allyson Ettinger, Amanda Bertsch, Bailey Kuehl, David Graham, David Heineman, Dirk Groeneveld, Faeze Brahman, Finbarr Timbers, Hamish Ivison, Jacob Morrison, Jake Poznanski, Kyle Lo, Luca Soldaini, Matt Jordan, Mayee Chen, Michael Noukhovitch, Nathan Lambert, Pete Walsh, Pradeep Dasigi, and 48 others. 2025.
\newblock \href {https://arxiv.org/abs/2512.13961} {{Olmo 3}}.
\newblock \emph{Preprint}, arXiv:2512.13961.

\bibitem[{Fang et~al.(2025)Fang, Fu, Zhao, and Wang}]{respec2025arxiv}
Min Fang, Zhihui Fu, Qibin Zhao, and Jun Wang. 2025.
\newblock \href {https://arxiv.org/abs/2511.01282} {{When, What, and How: Rethinking Retrieval-Enhanced Speculative Decoding}}.
\newblock \emph{Preprint}, arXiv:2511.01282.

\bibitem[{Grattafiori et~al.(2024)Grattafiori, Dubey, Jauhri, Pandey, Kadian, Al-Dahle, Letman, Mathur, Schelten, Vaughan, Yang, Fan, Goyal, Hartshorn, Yang, Mitra, Sravankumar, Korenev, Hinsvark, Rao, Zhang, Rodriguez, Gregerson, Spataru, Roziere, Biron, Tang, Chern, Caucheteux, Nayak, Bi, Marra, McConnell, Keller, Touret, Wu, Wong, Ferrer, Nikolaidis, Allonsius, Song, Pintz, Livshits, Wyatt, Esiobu, Choudhary, Mahajan, Garcia-Olano, Perino, Hupkes, Lakomkin, AlBadawy, Lobanova, Dinan, Smith, Radenovic, Guzmán, Zhang, Synnaeve, Lee, Anderson, Thattai, Nail, Mialon, Pang, Cucurell, Nguyen, Korevaar, Xu, Touvron, Zarov, Ibarra, Kloumann, Misra, Evtimov, Zhang, Copet, Lee, Geffert, Vranes, Park, Mahadeokar, Shah, van~der Linde, Billock, Hong, Lee, Fu, Chi, Huang, Liu, Wang, Yu, Bitton, Spisak, Park, Rocca, Johnstun, Saxe, Jia, Alwala, Prasad, Upasani, Plawiak, Li, Heafield, Stone, El-Arini, Iyer, Malik, Chiu, Bhalla, Lakhotia, Rantala-Yeary, van~der Maaten, Chen, Tan, Jenkins, Martin, Madaan, Malo, Blecher,
  Landzaat, de~Oliveira, Muzzi, Pasupuleti, Singh, Paluri, Kardas, Tsimpoukelli, Oldham, Rita, Pavlova, Kambadur, Lewis, Si, Singh, Hassan, Goyal, Torabi, Bashlykov, Bogoychev, Chatterji, Zhang, Duchenne, Çelebi, Alrassy, Zhang, Li, Vasic, Weng, Bhargava, Dubal, Krishnan, Koura, Xu, He, Dong, Srinivasan, Ganapathy, Calderer, Cabral, Stojnic, Raileanu, Maheswari, Girdhar, Patel, Sauvestre, Polidoro, Sumbaly, Taylor, Silva, Hou, Wang, Hosseini, Chennabasappa, Singh, Bell, Kim, Edunov, Nie, Narang, Raparthy, Shen, Wan, Bhosale, Zhang, Vandenhende, Batra, Whitman, Sootla, Collot, Gururangan, Borodinsky, Herman, Fowler, Sheasha, Georgiou, Scialom, Speckbacher, Mihaylov, Xiao, Karn, Goswami, Gupta, Ramanathan, Kerkez, Gonguet, Do, Vogeti, Albiero, Petrovic, Chu, Xiong, Fu, Meers, Martinet, Wang, Wang, Tan, Xia, Xie, Jia, Wang, Goldschlag, Gaur, Babaei, Wen, Song, Zhang, Li, Mao, Coudert, Yan, Chen, Papakipos, Singh, Srivastava, Jain, Kelsey, Shajnfeld, Gangidi, Victoria, Goldstand, Menon, Sharma, Boesenberg,
  Baevski, Feinstein, Kallet, Sangani, Teo, Yunus, Lupu, Alvarado, Caples, Gu, Ho, Poulton, Ryan, Ramchandani, Dong, Franco, Goyal, Saraf, Chowdhury, Gabriel, Bharambe, Eisenman, Yazdan, James, Maurer, Leonhardi, Huang, Loyd, Paola, Paranjape, Liu, Wu, Ni, Hancock, Wasti, Spence, Stojkovic, Gamido, Montalvo, Parker, Burton, Mejia, Liu, Wang, Kim, Zhou, Hu, Chu, Cai, Tindal, Feichtenhofer, Gao, Civin, Beaty, Kreymer, Li, Adkins, Xu, Testuggine, David, Parikh, Liskovich, Foss, Wang, Le, Holland, Dowling, Jamil, Montgomery, Presani, Hahn, Wood, Le, Brinkman, Arcaute, Dunbar, Smothers, Sun, Kreuk, Tian, Kokkinos, Ozgenel, Caggioni, Kanayet, Seide, Florez, Schwarz, Badeer, Swee, Halpern, Herman, Sizov, Guangyi, Zhang, Lakshminarayanan, Inan, Shojanazeri, Zou, Wang, Zha, Habeeb, Rudolph, Suk, Aspegren, Goldman, Zhan, Damlaj, Molybog, Tufanov, Leontiadis, Veliche, Gat, Weissman, Geboski, Kohli, Lam, Asher, Gaya, Marcus, Tang, Chan, Zhen, Reizenstein, Teboul, Zhong, Jin, Yang, Cummings, Carvill, Shepard, McPhie,
  Torres, Ginsburg, Wang, Wu, U, Saxena, Khandelwal, Zand, Matosich, Veeraraghavan, Michelena, Li, Jagadeesh, Huang, Chawla, Huang, Chen, Garg, A, Silva, Bell, Zhang, Guo, Yu, Moshkovich, Wehrstedt, Khabsa, Avalani, Bhatt, Mankus, Hasson, Lennie, Reso, Groshev, Naumov, Lathi, Keneally, Liu, Seltzer, Valko, Restrepo, Patel, Vyatskov, Samvelyan, Clark, Macey, Wang, Hermoso, Metanat, Rastegari, Bansal, Santhanam, Parks, White, Bawa, Singhal, Egebo, Usunier, Mehta, Laptev, Dong, Cheng, Chernoguz, Hart, Salpekar, Kalinli, Kent, Parekh, Saab, Balaji, Rittner, Bontrager, Roux, Dollar, Zvyagina, Ratanchandani, Yuvraj, Liang, Alao, Rodriguez, Ayub, Murthy, Nayani, Mitra, Parthasarathy, Li, Hogan, Battey, Wang, Howes, Rinott, Mehta, Siby, Bondu, Datta, Chugh, Hunt, Dhillon, Sidorov, Pan, Mahajan, Verma, Yamamoto, Ramaswamy, Lindsay, Lindsay, Feng, Lin, Zha, Patil, Shankar, Zhang, Zhang, Wang, Agarwal, Sajuyigbe, Chintala, Max, Chen, Kehoe, Satterfield, Govindaprasad, Gupta, Deng, Cho, Virk, Subramanian, Choudhury,
  Goldman, Remez, Glaser, Best, Koehler, Robinson, Li, Zhang, Matthews, Chou, Shaked, Vontimitta, Ajayi, Montanez, Mohan, Kumar, Mangla, Ionescu, Poenaru, Mihailescu, Ivanov, Li, Wang, Jiang, Bouaziz, Constable, Tang, Wu, Wang, Wu, Gao, Kleinman, Chen, Hu, Jia, Qi, Li, Zhang, Zhang, Adi, Nam, Yu, Wang, Zhao, Hao, Qian, Li, He, Rait, DeVito, Rosnbrick, Wen, Yang, Zhao, and Ma}]{grattafiori2024llama3herdmodels}
Aaron Grattafiori, Abhimanyu Dubey, Abhinav Jauhri, Abhinav Pandey, Abhishek Kadian, Ahmad Al-Dahle, Aiesha Letman, Akhil Mathur, Alan Schelten, Alex Vaughan, Amy Yang, Angela Fan, Anirudh Goyal, Anthony Hartshorn, Aobo Yang, Archi Mitra, Archie Sravankumar, Artem Korenev, Arthur Hinsvark, and 542 others. 2024.
\newblock \href {https://arxiv.org/abs/2407.21783} {{The Llama 3 Herd of Models}}.
\newblock \emph{Preprint}, arXiv:2407.21783.

\bibitem[{Hendrycks and Gimpel(2016)}]{hendrycks-gimpel-2016-gaussian}
Dan Hendrycks and Kevin Gimpel. 2016.
\newblock \href {https://arxiv.org/abs/1606.08415} {{Gaussian Error Linear Units (GELUs)}}.
\newblock \emph{Preprint}, arXiv:1606.08415.

\bibitem[{Jain et~al.(2025)Jain, Han, Gu, Li, Yan, Zhang, Wang, Solar-Lezama, Sen, and Stoica}]{livecodebench2024}
Naman Jain, King Han, Alex Gu, Wen-Ding Li, Fanjia Yan, Tianjun Zhang, Sida Wang, Armando Solar-Lezama, Koushik Sen, and Ion Stoica. 2025.
\newblock \href {https://openreview.net/forum?id=chfJJYC3iL} {{LiveCodeBench}: Holistic and contamination free evaluation of large language models for code}.
\newblock In \emph{The Thirteenth International Conference on Learning Representations}.

\bibitem[{Jimenez et~al.(2024)Jimenez, Yang, Wettig, Yao, Pei, Press, and Narasimhan}]{swebench2024}
Carlos~E Jimenez, John Yang, Alexander Wettig, Shunyu Yao, Kexin Pei, Ofir Press, and Karthik~R Narasimhan. 2024.
\newblock \href {https://openreview.net/forum?id=VTF8yNQM66} {{{SWE}-bench: Can Language Models Resolve Real-world Github Issues?}}
\newblock In \emph{The Twelfth International Conference on Learning Representations}.

\bibitem[{Kim et~al.(2026{\natexlab{a}})Kim, Shin, Chung, and Rhu}]{kim-etal-2026-cost}
Jiin Kim, Byeongjun Shin, Jinha Chung, and Minsoo Rhu. 2026{\natexlab{a}}.
\newblock \href {https://doi.org/10.1109/HPCA68181.2026.11408569} {{The Cost of Dynamic Reasoning: Demystifying AI Agents and Test-Time Scaling from an AI Infrastructure Perspective}}.
\newblock In \emph{2026 IEEE International Symposium on High Performance Computer Architecture (HPCA)}, pages 1--16.

\bibitem[{Kim et~al.(2026{\natexlab{b}})Kim, Jung, and Yun}]{kim-etal-2026-multi}
Taehyeon Kim, Hojung Jung, and Se-Young Yun. 2026{\natexlab{b}}.
\newblock \href {https://doi.org/10.18653/v1/2026.findings-acl.1629} {Multi-drafter speculative decoding with alignment feedback}.
\newblock In \emph{Findings of the {A}ssociation for {C}omputational {L}inguistics: {ACL} 2026}, pages 32532--32573, San Diego, California, United States. Association for Computational Linguistics.

\bibitem[{Leviathan et~al.(2023)Leviathan, Kalman, and Matias}]{sd2023icml}
Yaniv Leviathan, Matan Kalman, and Yossi Matias. 2023.
\newblock \href {https://proceedings.mlr.press/v202/leviathan23a.html} {{Fast Inference from Transformers via Speculative Decoding}}.
\newblock In \emph{Proceedings of the 40th International Conference on Machine Learning}, volume 202 of \emph{Proceedings of Machine Learning Research}, pages 19274--19286. PMLR.

\bibitem[{Li et~al.(2025)Li, Wei, Zhang, and Zhang}]{eagle3_2025neurips}
Yuhui Li, Fangyun Wei, Chao Zhang, and Hongyang Zhang. 2025.
\newblock \href {https://proceedings.neurips.cc/paper_files/paper/2025/file/c7b5a35ea98b62512a869c19ea7b03cb-Paper-Conference.pdf} {{EAGLE-3: Scaling up Inference Acceleration of Large Language Models via Training-Time Test}}.
\newblock In \emph{Advances in Neural Information Processing Systems}, volume~38, pages 136737--136756. Curran Associates, Inc.

\bibitem[{Lightman et~al.(2024)Lightman, Kosaraju, Burda, Edwards, Baker, Lee, Leike, Schulman, Sutskever, and Cobbe}]{math500_2023}
Hunter Lightman, Vineet Kosaraju, Yuri Burda, Harrison Edwards, Bowen Baker, Teddy Lee, Jan Leike, John Schulman, Ilya Sutskever, and Karl Cobbe. 2024.
\newblock \href {https://openreview.net/forum?id=v8L0pN6EOi} {{Let's Verify Step by Step}}.
\newblock In \emph{The Twelfth International Conference on Learning Representations}.

\bibitem[{Liu et~al.(2025)Liu, Mei, Lin, Xue, Wang, Xu, Wu, Zhang, Lin, Dong, Lu, Zhao, Deng, Xu, Ruan, Dai, Guo, Yang, Chen, Li, Zhou, Lin, Dai, Hao, Chen, Li, Zhang, Xu, Li, Liang, Wei, Zhang, Luo, Ji, Ding, Tang, Cao, Gao, Qu, Zeng, Huang, Li, Xu, Hu, Chen, Xiang, Yuan, Cheng, Zhu, Ran, Jiang, Qiu, Li, Song, Dong, Gao, Guan, Huang, Zhou, Huang, Yu, Wang, Zhang, Wang, Zhao, Yin, Guo, Luo, Ma, Wang, Zhang, Di, Xu, Zhang, Zhang, Tang, Zhou, Huang, Cong, Wang, Wang, Zhu, Li, Chen, Du, Xu, Ge, Zhang, Pan, Wang, Yin, Xu, Shen, Zhang, Liu, Lu, Zhou, Chen, Cai, Chen, Hu, Liu, Hu, Ma, Wang, Yu, Zhou, Pan, Zhou, Ni, Yun, Pei, Ye, Yue, Zeng, Liu, Liang, Pang, Luo, Gao, Zhang, Gao, Wang, Bi, Liu, Wang, Chen, Zhang, Nie, Cheng, Liu, Xie, Liu, Yu, Li, Yang, Li, Chen, Su, Pan, Lin, Fu, Wang, Zhang, Xu, Ma, Li, Li, Zhao, Sun, Wang, Qian, Yu, Zhang, Ding, Shi, Xiong, He, Zhou, Zhong, Piao, Wang, Chen, Tan, Wei, Ma, Liu, Yang, Guo, Wu, Wu, Cheng, Ou, Xu, Wang, Gong, Wu, Zou, Li, Xiong, Luo, You, Liu, Zhou, Wu, Ren, Zhao,
  Ren, Sha, Fu, Xu, Xie, Zhang, Hao, Gou, Ma, Yan, Shao, Huang, Wu, Li, Zhang, Xu, Wang, Gu, Zhu, Li, Zhang, Xie, Gao, Pan, Yao, Feng, Li, Cai, Ni, Xu, Li, Tian, Chen, Jin, Li, Zhou, Sun, Li, Jin, Shen, Chen, Song, Zhou, Zhu, Huang, Li, Zheng, Zhu, Ma, Huang, Xu, Zhang, Ji, Liang, Guo, Chen, Xia, Wang, Li, Zhang, Chen, Sun, Wu, Ye, Wang, Xiao, An, Wang, Sun, Wang, Tang, Zha, Zhang, Ju, Zhang, and Qu}]{liu-etal-2025-deepseekv32}
Aixin Liu, Aoxue Mei, Bangcai Lin, Bing Xue, Bingxuan Wang, Bingzheng Xu, Bochao Wu, Bowei Zhang, Chaofan Lin, Chen Dong, Chengda Lu, Chenggang Zhao, Chengqi Deng, Chenhao Xu, Chong Ruan, Damai Dai, Daya Guo, Dejian Yang, Deli Chen, and 244 others. 2025.
\newblock \href {https://arxiv.org/abs/2512.02556} {{DeepSeek-V3.2: Pushing the Frontier of Open Large Language Models}}.
\newblock \emph{Preprint}, arXiv:2512.02556.

\bibitem[{Liu et~al.(2026{\natexlab{a}})Liu, Huang, Jia, Park, and Wang}]{hedgespec2026iclr}
Hongyi Liu, Jiaji Huang, Zhen Jia, Youngsuk Park, and Yu-Xiang Wang. 2026{\natexlab{a}}.
\newblock \href {https://openreview.net/forum?id=JMmljf895g} {{Not-a-Bandit: Provably No-Regret Drafter Selection in Speculative Decoding for {LLM}s}}.
\newblock In \emph{The Fourteenth International Conference on Learning Representations}.

\bibitem[{Liu et~al.(2026{\natexlab{b}})Liu, Lv, Shen, Sun, and Sun}]{liu2026talon}
Tianyu Liu, Qitan Lv, Yuhao Shen, Xiao Sun, and Xiaoyan Sun. 2026{\natexlab{b}}.
\newblock \href {https://arxiv.org/abs/2601.07353} {{{TALON}: Confidence-Aware Speculative Decoding with Adaptive Token Trees}}.
\newblock \emph{Preprint}, arXiv:2601.07353.

\bibitem[{Liu et~al.(2024)Liu, Yu, Zhang, Xu, Lei, Lai, Gu, Ding, Men, Yang, Zhang, Deng, Zeng, Du, Zhang, Shen, Zhang, Su, Sun, Huang, Dong, and Tang}]{agentbench2023}
Xiao Liu, Hao Yu, Hanchen Zhang, Yifan Xu, Xuanyu Lei, Hanyu Lai, Yu~Gu, Hangliang Ding, Kaiwen Men, Kejuan Yang, Shudan Zhang, Xiang Deng, Aohan Zeng, Zhengxiao Du, Chenhui Zhang, Sheng Shen, Tianjun Zhang, Yu~Su, Huan Sun, and 3 others. 2024.
\newblock \href {https://openreview.net/forum?id=zAdUB0aCTQ} {{AgentBench}: Evaluating {LLM}s as agents}.
\newblock In \emph{The Twelfth International Conference on Learning Representations}.

\bibitem[{Loshchilov and Hutter(2019)}]{loshchilov-hutter-2018-decoupled}
Ilya Loshchilov and Frank Hutter. 2019.
\newblock \href {https://openreview.net/forum?id=Bkg6RiCqY7} {Decoupled weight decay regularization}.
\newblock In \emph{International Conference on Learning Representations}.

\bibitem[{Ma et~al.(2025)Ma, Gim, and Zhong}]{ma-etal-2025-cacheback}
Zhiyao Ma, In~Gim, and Lin Zhong. 2025.
\newblock \href {https://doi.org/10.18653/v1/2025.emnlp-main.1581} {Cacheback: Speculative decoding with nothing but cache}.
\newblock In \emph{Proceedings of the 2025 Conference on Empirical Methods in Natural Language Processing}, pages 31079--31084, Suzhou, China. Association for Computational Linguistics.

\bibitem[{Mitra et~al.(2024)Mitra, Corro, Zheng, Mahajan, Rouhana, Codas, Lu, ge~Chen, Vrousgos, Rosset, Silva, Khanpour, Lara, and Awadallah}]{agentinstruct2024}
Arindam Mitra, Luciano~Del Corro, Guoqing Zheng, Shweti Mahajan, Dany Rouhana, Andres Codas, Yadong Lu, Wei ge~Chen, Olga Vrousgos, Corby Rosset, Fillipe Silva, Hamed Khanpour, Yash Lara, and Ahmed Awadallah. 2024.
\newblock \href {https://arxiv.org/abs/2407.03502} {{AgentInstruct}: Toward generative teaching with agentic flows}.
\newblock \emph{Preprint}, arXiv:2407.03502.

\bibitem[{Oliaro et~al.(2025)Oliaro, Jia, Campos, and Qiao}]{suffixdecoding2025neurips}
Gabriele Oliaro, Zhihao Jia, Daniel Campos, and Aurick Qiao. 2025.
\newblock \href {https://proceedings.neurips.cc/paper_files/paper/2025/file/b7aea253ab34a773967f1e4cdea9e4fb-Paper-Conference.pdf} {{SuffixDecoding: Extreme Speculative Decoding for Emerging {AI} Applications}}.
\newblock In \emph{Advances in Neural Information Processing Systems}, volume~38, pages 126326--126354. Curran Associates, Inc.

\bibitem[{Quan et~al.(2025)Quan, Feng, Hao, Jiang, Zhang, and Wang}]{quan-etal-2025-rasd}
Guofeng Quan, Wenfeng Feng, Chuzhan Hao, Guochao Jiang, Yuewei Zhang, and Hao~Henry Wang. 2025.
\newblock \href {https://doi.org/10.18653/v1/2025.findings-acl.320} {{RASD}: Retrieval-augmented speculative decoding}.
\newblock In \emph{Findings of the Association for Computational Linguistics: ACL 2025}, pages 6167--6177, Vienna, Austria. Association for Computational Linguistics.

\bibitem[{Ringel and Romano(2026)}]{ddtree2026arxiv}
Liran Ringel and Yaniv Romano. 2026.
\newblock \href {https://arxiv.org/abs/2604.12989} {{Accelerating Speculative Decoding with Block Diffusion Draft Trees}}.
\newblock \emph{Preprint}, arXiv:2604.12989.

\bibitem[{Taori et~al.(2023)Taori, Gulrajani, Zhang, Dubois, Li, Guestrin, Liang, and Hashimoto}]{alpaca2023}
Rohan Taori, Ishaan Gulrajani, Tianyi Zhang, Yann Dubois, Xuechen Li, Carlos Guestrin, Percy Liang, and Tatsunori~B. Hashimoto. 2023.
\newblock {Stanford Alpaca: An Instruction-following LLaMA model}.
\newblock \url{https://github.com/tatsu-lab/stanford_alpaca}.

\bibitem[{Wang et~al.(2025)Wang, Su, Li, Xia, Ye, Duan, Wang, and Zhang}]{wang-etal-2025-opt}
Jikai Wang, Yi~Su, Juntao Li, Qingrong Xia, Zi~Ye, Xinyu Duan, Zhefeng Wang, and Min Zhang. 2025.
\newblock \href {https://doi.org/10.1162/tacl_a_00735} {{OPT}-tree: Speculative decoding with adaptive draft tree structure}.
\newblock \emph{Transactions of the Association for Computational Linguistics}, 13:188--199.

\bibitem[{Williams et~al.(2026)Williams, Kwon, Li, Kouris, and Venieris}]{williams-etal-2026-speculative}
Miles Williams, Young~D. Kwon, Rui Li, Alexandros Kouris, and Stylianos~I. Venieris. 2026.
\newblock \href {https://doi.org/10.18653/v1/2026.findings-acl.2000} {Speculative decoding with a speculative vocabulary}.
\newblock In \emph{Findings of the {A}ssociation for {C}omputational {L}inguistics: {ACL} 2026}, pages 40240--40254, San Diego, California, United States. Association for Computational Linguistics.

\bibitem[{Wolf et~al.(2020)Wolf, Debut, Sanh, Chaumond, Delangue, Moi, Cistac, Rault, Louf, Funtowicz, Davison, Shleifer, von Platen, Ma, Jernite, Plu, Xu, Le~Scao, Gugger, Drame, Lhoest, and Rush}]{wolf-etal-2020-transformers}
Thomas Wolf, Lysandre Debut, Victor Sanh, Julien Chaumond, Clement Delangue, Anthony Moi, Pierric Cistac, Tim Rault, Remi Louf, Morgan Funtowicz, Joe Davison, Sam Shleifer, Patrick von Platen, Clara Ma, Yacine Jernite, Julien Plu, Canwen Xu, Teven Le~Scao, Sylvain Gugger, and 3 others. 2020.
\newblock \href {https://doi.org/10.18653/v1/2020.emnlp-demos.6} {Transformers: State-of-the-art natural language processing}.
\newblock In \emph{Proceedings of the 2020 Conference on Empirical Methods in Natural Language Processing: System Demonstrations}, pages 38--45, Online. Association for Computational Linguistics.

\bibitem[{Yang et~al.(2025)Yang, Li, Yang, Zhang, Hui, Zheng, Yu, Gao, Huang, Lv, Zheng, Liu, Zhou, Huang, Hu, Ge, Wei, Lin, Tang, Yang, Tu, Zhang, Yang, Yang, Zhou, Zhou, Lin, Dang, Bao, Yang, Yu, Deng, Li, Xue, Li, Zhang, Wang, Zhu, Men, Gao, Liu, Luo, Li, Tang, Yin, Ren, Wang, Zhang, Ren, Fan, Su, Zhang, Zhang, Wan, Liu, Wang, Cui, Zhang, Zhou, and Qiu}]{yang-etal-2025-qwen3}
An~Yang, Anfeng Li, Baosong Yang, Beichen Zhang, Binyuan Hui, Bo~Zheng, Bowen Yu, Chang Gao, Chengen Huang, Chenxu Lv, Chujie Zheng, Dayiheng Liu, Fan Zhou, Fei Huang, Feng Hu, Hao Ge, Haoran Wei, Huan Lin, Jialong Tang, and 41 others. 2025.
\newblock \href {https://arxiv.org/abs/2505.09388} {{Qwen3 Technical Report}}.
\newblock \emph{Preprint}, arXiv:2505.09388.

\bibitem[{Yao et~al.(2025)Yao, Shinn, Razavi, and Narasimhan}]{taubench2024}
Shunyu Yao, Noah Shinn, Pedram Razavi, and Karthik~R Narasimhan. 2025.
\newblock \href {https://openreview.net/forum?id=roNSXZpUDN} {{$\tau$-bench: A Benchmark for Tool-Agent-User Interaction in Real-World Domains}}.
\newblock In \emph{The Thirteenth International Conference on Learning Representations}.

\bibitem[{Zhao et~al.(2024)Zhao, Ren, Hessel, Cardie, Choi, and Deng}]{wildchat2024}
Wenting Zhao, Xiang Ren, Jack Hessel, Claire Cardie, Yejin Choi, and Yuntian Deng. 2024.
\newblock \href {https://openreview.net/forum?id=Bl8u7ZRlbM} {{WildChat}: {1M} {ChatGPT} interaction logs in the wild}.
\newblock In \emph{The Twelfth International Conference on Learning Representations}.

\bibitem[{Zheng et~al.(2023)Zheng, Chiang, Sheng, Zhuang, Wu, Zhuang, Lin, Li, Li, Xing, Zhang, Gonzalez, and Stoica}]{zheng2023judging}
Lianmin Zheng, Wei-Lin Chiang, Ying Sheng, Siyuan Zhuang, Zhanghao Wu, Yonghao Zhuang, Zi~Lin, Zhuohan Li, Dacheng Li, Eric Xing, Hao Zhang, Joseph Gonzalez, and Ion Stoica. 2023.
\newblock \href {https://doi.org/10.52202/075280-2020} {{Judging LLM-as-a-Judge with MT-Bench and Chatbot Arena}}.
\newblock In \emph{Advances in Neural Information Processing Systems}, volume~36, pages 46595--46623. Curran Associates, Inc.

\end{thebibliography}

\clearpage
\appendix

\section{Detailed Experimental Setup}\label{app:experimental_setup}

\paragraph{Neural Router Architecture.}
\label{sec:neural_router_architecture}

The neural policy is a multilayer perceptron that maps the target model's final hidden state to a scalar output. Given a final hidden state $h_t \in \mathbb{R}^d$ where $d$ is the target model dimensionality, the router first projects $h_t$ to an intermediate dimension $m = \frac{d}{8}$. This intermediary representation is then supplied to a Gaussian Error Linear Unit (GELU) activation~\citep{hendrycks-gimpel-2016-gaussian}, followed by dropout. Finally, the intermediary representation is projected to a scalar output $y_t \in \mathbb{R}$, used for the router policy.

\paragraph{Drafter Hyperparameters.}
For the autoregressive drafter, we follow the original EAGLE-3 paper~\citep{eagle3_2025neurips}, using a draft-tree depth of $8$, top-$k$ of $10$, and $60$ total draft tokens. For the diffusion drafter, DFlash~\citep{dflash2026icml} uses a block size of $16$ for the Qwen3 models (4B and 8B) and $10$ for Llama3.1-8B, the default block size for each drafter model. \ours routes between these two drafters under identical configurations, so each standalone baseline is matched to its counterpart inside \ours. Table~\ref{tab:app:drafter_hparams} summarises the hyperparameter settings.

\paragraph{Draft Models.}
For our experiments, we use the Qwen3 EAGLE-3 draft models from \citet{williams-etal-2026-speculative}, which outperform previously released checkpoints; all other draft models are the official releases.

\paragraph{Datasets.}
We report statistics for our evaluation datasets in Table~\ref{tab:app:datasets} and group datasets based on their task category.

\paragraph{Training Duration.}
We have compiled the neural router training cost (Table~\ref{tab:app:router_training_cost}). We observe that the training cost is very affordable, consistently taking less than one hour on one H100 GPU.

\paragraph{Software.}
In all experiments, we use the model and tokeniser implementations for \textit{Qwen3} and \textit{Llama3.1} provided by the Hugging Face Transformers framework \citep{wolf-etal-2020-transformers}, v\texttt{4.57.1}.

\begin{table}[t]
\small
\centering
\begin{tabular}{ll}
\toprule
Method & Hyperparameters \\
\midrule
EAGLE-3  & Depth $8$, Top-$k$ $10$, Draft tokens $60$. \\
DFlash   & Block size $16$ (Qwen3), $10$ (Llama-3.1-8B). \\
\ours    & Identical to the above. \\
\bottomrule
\end{tabular}
\caption{Drafter hyperparameter settings for EAGLE-3, DFlash, and \ours.}
\label{tab:app:drafter_hparams}
\end{table}

\section{Additional Results}\label{app:results}

In this section, we present the full results for all the models employed in our study, in addition to the results presented in Section~\ref{sec:results}. 

Tables~\ref{tab:app:accept_length} and~\ref{tab:app:speedup} present per-dataset acceptance length (AL) and speedup ($\times$) results for three target models across 15 datasets from four task categories.

\begin{table}[t]
  \small
  \centering
  \begin{tabular}{ll c}
    \toprule
    \textbf{Category} & \textbf{Dataset/Task} & \textbf{Examples} \\
    \midrule
\multirow{4}{*}{Chat} & MT-Bench & 80 \\
& Alpaca & 128 \\
& UltraChat & 80 \\
& WildChat & 80 \\
    \midrule
\multirow{4}{*}{Math} & GSM8K & 128 \\
& MATH-500 & 128 \\
& AIME24 & 30 \\
& AIME25 & 30 \\
\midrule
\multirow{3}{*}{Coding} & HumanEval & 164 \\
& MBPP & 128 \\
& LiveCodeBench & 128 \\
\midrule
\multirow{4}{*}{Agentic} & SWE-Bench & 128 \\
& Tau-Bench & 162 \\
& AgentBench & 103 \\
& AgentInstruct & 80 \\
\midrule
\textbf{Total} & & 1577 \\
    \bottomrule
  \end{tabular}
  \caption{Number of examples in each dataset of our evaluation suite.}
  \label{tab:app:datasets}
\end{table}

\begin{table}[t]
\small
\centering
\begin{tabular}{lc}
\toprule
\textbf{Model} & \textbf{Duration (mm:ss)} \\
\midrule
Qwen3-4B    & 20:23 \\
Qwen3-8B    & 35:26 \\
Llama3.1-8B & 24:44 \\
\bottomrule
\end{tabular}
\caption{Training duration for each neural router on a single NVIDIA H100 GPU.}
\label{tab:app:router_training_cost}
\end{table}

\begin{table*}[t]
  \scriptsize
  \centering
  \begin{tabular}{lll ccccccc}
    \toprule
  & & & \multicolumn{7}{c}{\textbf{Method}} \\
  \cmidrule(lr){4-10}
  \textbf{Model} & \textbf{Category} & \textbf{Dataset}
  & \textbf{EAGLE-3}
  & \textbf{DFlash}
  & \shortstack{\textbf{Oracle}\\\textbf{-Task}}
  & \shortstack{\textbf{Oracle}\\\textbf{-Prompt}}
  & \shortstack{\textbf{Oracle}\\\textbf{-Token}}
  & \shortstack{\textbf{\ours}\\\textbf{-Entropy}}
  & \shortstack{\textbf{\ours}\\\textbf{-Neural}} \\
    \midrule
\multirow{15}{*}{Qwen3-4B} & \multirow{4}{*}{Chat} & MT-Bench & 5.14	&	4.30	&	5.14	&	5.44	&	6.08	&	5.46	&	5.63 \\
& & Alpaca & 5.06	&	2.90	&	5.06	&	5.01	&	5.31	&	4.90	&	5.12 \\
& & UltraChat & 5.79	&	2.93	&	5.79	&	5.79	&	5.95	&	5.57	&	5.81 \\
& & WildChat & 4.83	&	3.58	&	4.83	&	4.91	&	5.39	&	4.82	&	5.07 \\
    \cmidrule{2-10}
& \multirow{4}{*}{Math} & GSM8K & 5.90	&	6.33	&	6.33	&	6.34	&	7.80	&	6.57	&	7.07 \\
& & MATH-500 & 5.18	&	7.88	&	7.88	&	7.89	&	8.78	&	7.90	&	8.02 \\
& & AIME24 & 5.44	&	7.58	&	7.58	&	7.58	&	8.80	&	7.77	&	7.91 \\
& & AIME25 & 5.25	&	7.09	&	7.09	&	7.09	&	8.06	&	7.21	&	7.63 \\
\cmidrule{2-10}
& \multirow{3}{*}{Coding} & HumanEval & 5.56	&	6.39	&	6.39	&	6.42	&	7.93	&	6.85	&	7.13 \\
& & MBPP & 5.73	&	6.08	&	6.08	&	6.13	&	7.73	&	6.43	&	6.76 \\
& & LiveCodeBench & 4.45	&	6.87	&	6.87	&	6.93	&	7.70	&	6.91	&	7.01 \\
\cmidrule{2-10}
& \multirow{4}{*}{Agentic} & SWE-Bench & 4.04	&	3.55	&	4.04	&	3.88	&	4.95	&	4.32	&	4.51 \\
& & Tau-Bench & 3.13	&	2.79	&	3.13	&	3.04	&	3.68	&	3.23	&	3.29 \\
& & AgentBench & 4.57	&	3.82	&	4.57	&	4.59	&	5.34	&	4.72	&	4.94 \\
& & AgentInstruct & 4.21	&	4.07	&	4.21	&	4.54	&	5.28	&	4.59	&	4.87 \\
\midrule
\multirow{15}{*}{Qwen3-8B} & \multirow{4}{*}{Chat} & MT-Bench & 5.17	&	4.26	&	5.17	&	5.46	&	6.13	&	5.41	&	5.68 \\
& & Alpaca & 5.26	&	3.04	&	5.26	&	5.29	&	5.50	&	5.10	&	5.34 \\
& & UltraChat & 5.82	&	2.84	&	5.82	&	5.82	&	5.93	&	5.62	&	5.84 \\
& & WildChat & 4.70	&	3.47	&	4.70	&	4.79	&	5.30	&	4.83	&	5.02 \\
    \cmidrule{2-10}
& \multirow{4}{*}{Math} & GSM8K & 6.21	&	6.48	&	6.48	&	6.56	&	8.21	&	6.91	&	7.35 \\
& & MATH-500 & 5.53	&	7.91	&	7.91	&	7.91	&	9.03	&	7.98	&	8.11 \\
& & AIME24 & 5.29	&	7.82	&	7.82	&	7.82	&	8.79	&	7.81	&	7.93 \\
& & AIME25 & 5.28	&	7.29	&	7.29	&	7.29	&	8.20	&	7.46	&	7.49 \\
\cmidrule{2-10}
& \multirow{3}{*}{Coding} & HumanEval & 5.65	&	6.52	&	6.52	&	6.52	&	8.22	&	7.08	&	7.42 \\
& & MBPP & 5.78	&	5.97	&	5.97	&	6.10	&	7.76	&	6.47	&	6.83 \\
& & LiveCodeBench & 4.44	&	7.20	&	7.20	&	7.27	&	7.97	&	7.35	&	7.34 \\
\cmidrule{2-10}
& \multirow{4}{*}{Agentic} & SWE-Bench & 4.07	&	3.57	&	4.07	&	3.81	&	5.02	&	4.44	&	4.60 \\
& & Tau-Bench & 4.44	&	4.59	&	4.59	&	4.84	&	5.24	&	4.88	&	4.82 \\
& & AgentBench & 4.48	&	3.74	&	4.48	&	4.44	&	5.06	&	4.63	&	4.80 \\
& & AgentInstruct & 4.67	&	4.58	&	4.67	&	5.03	&	5.72	&	5.17	&	5.38 \\
\midrule
\multirow{15}{*}{Llama3.1-8B} & \multirow{4}{*}{Chat} & MT-Bench & 5.19	&	3.97	&	5.19	&	4.54	&	5.59	&	5.20	&	5.26 \\
& & Alpaca & 4.96	&	3.37	&	4.96	&	4.37	&	5.18	&	4.93	&	4.96 \\
& & UltraChat & 5.38	&	3.96	&	5.38	&	4.95	&	5.73	&	5.33	&	5.39 \\
& & WildChat & 4.42	&	3.16	&	4.42	&	3.91	&	4.71	&	4.41	&	4.47 \\
    \cmidrule{2-10}
& \multirow{4}{*}{Math} & GSM8K & 5.61	&	4.31	&	5.61	&	4.80	&	6.05	&	5.57	&	5.63 \\
& & MATH-500 & 4.97	&	4.27	&	4.97	&	4.64	&	5.56	&	4.90	&	5.08 \\
& & AIME24 & 5.20	&	4.70	&	5.20	&	4.97	&	5.85	&	5.31	&	5.37 \\
& & AIME25 & 5.23	&	4.66	&	5.23	&	4.98	&	6.20	&	5.58	&	5.83 \\
\cmidrule{2-10}
& \multirow{3}{*}{Coding} & HumanEval & 6.05	&	5.01	&	6.05	&	5.19	&	6.76	&	6.01	&	6.28 \\
& & MBPP & 6.17	&	5.00	&	6.17	&	5.44	&	6.75	&	6.11	&	6.35 \\
& & LiveCodeBench & 4.63	&	3.95	&	4.63	&	4.14	&	5.18	&	4.61	&	4.73 \\
\cmidrule{2-10}
& \multirow{4}{*}{Agentic} & SWE-Bench & 4.38	&	3.81	&	4.38	&	3.90	&	5.16	&	4.47	&	4.75 \\
& & Tau-Bench & 4.59	&	2.80	&	4.59	&	4.44	&	4.79	&	4.59	&	4.64 \\
& & AgentBench & 4.12	&	3.40	&	4.12	&	3.93	&	4.54	&	4.12	&	4.27 \\
& & AgentInstruct & 4.02	&	3.12	&	4.02	&	3.44	&	4.41	&	4.02	&	4.12 \\
    \bottomrule
  \end{tabular}
  \caption{Per-dataset acceptance length (AL) results for three target models across 15 datasets from four task categories.}
  \label{tab:app:accept_length}
\end{table*}

\begin{table*}[t]
  \scriptsize
  \centering
  \begin{tabular}{lll cccccc}
    \toprule
  & & & \multicolumn{6}{c}{\textbf{Method}} \\
  \cmidrule(lr){4-9}
  \textbf{Model} & \textbf{Category} & \textbf{Dataset}
  & \textbf{EAGLE-3}
  & \textbf{DFlash}
  & \shortstack{\textbf{Oracle}\\\textbf{-Task}}
  & \shortstack{\textbf{Oracle}\\\textbf{-Prompt}}
  & \shortstack{\textbf{\ours}\\\textbf{-Entropy}}
  & \shortstack{\textbf{\ours}\\\textbf{-Neural}} \\
    \midrule
\multirow{15}{*}{Qwen3-4B} & \multirow{4}{*}{Chat} & MT-Bench & 2.70	&	2.44	&	2.70	&	2.98	&	2.96	&	2.91	\\
& & Alpaca & 2.45	&	1.86	&	2.45	&	2.51	&	2.39	&	2.54	\\
& & UltraChat & 2.95	&	1.89	&	2.95	&	2.93	&	2.96	&	3.11	\\
& & WildChat & 2.24	&	2.15	&	2.24	&	2.63	&	2.46	&	2.58	\\
    \cmidrule{2-9}
& \multirow{4}{*}{Math} & GSM8K & 3.25	&	4.30	&	4.30	&	4.62	&	4.38	&	4.29	\\
& & MATH-500 & 2.73	&	5.15	&	5.15	&	5.10	&	5.38	&	5.40	\\
& & AIME24 & 2.93	&	4.96	&	4.96	&	5.26	&	5.02	&	5.13	\\
& & AIME25 & 2.84	&	4.57	&	4.57	&	5.02	&	4.83	&	5.09	\\
\cmidrule{2-9}
& \multirow{3}{*}{Coding} & HumanEval & 3.11	&	4.23	&	4.23	&	4.65	&	4.57	&	4.68	\\
& & MBPP & 2.96	&	4.01	&	4.01	&	4.19	&	4.14	&	4.27	\\
& & LiveCodeBench & 2.31	&	4.62	&	4.62	&	4.51	&	4.66	&	4.75	\\
\cmidrule{2-9}
& \multirow{4}{*}{Agentic} & SWE-Bench & 2.25	&	2.60	&	2.60	&	2.61	&	2.78	&	2.73	\\
& & Tau-Bench & 1.44	&	1.72	&	1.72	&	1.76	&	1.77	&	1.76	\\
& & AgentBench & 2.27	&	2.62	&	2.62	&	2.66	&	2.57	&	2.60	\\
& & AgentInstruct & 2.12	&	2.43	&	2.43	&	2.62	&	2.50	&	2.59	\\
\midrule
\multirow{15}{*}{Qwen3-8B} & \multirow{4}{*}{Chat} & MT-Bench & 2.73	&	2.39	&	2.73	&	2.95	&	2.95	&	3.00	\\
& & Alpaca & 2.65	&	1.96	&	2.65	&	2.98	&	2.68	&	2.84	\\
& & UltraChat & 3.30	&	2.02	&	3.30	&	3.32	&	3.23	&	3.32	\\
& & WildChat & 2.22	&	2.12	&	2.22	&	2.54	&	2.51	&	2.54	\\
    \cmidrule{2-9}
& \multirow{4}{*}{Math} & GSM8K & 3.49	&	4.54	&	4.54	&	4.59	&	4.50	&	4.69	\\
& & MATH-500 & 3.03	&	5.38	&	5.38	&	5.47	&	5.46	&	5.39	\\
& & AIME24 & 3.07	&	5.99	&	5.99	&	5.45	&	5.75	&	5.74	\\
& & AIME25 & 2.92	&	5.00	&	5.00	&	5.28	&	5.39	&	5.36	\\
\cmidrule{2-9}
& \multirow{3}{*}{Coding} & HumanEval & 3.19	&	4.64	&	4.64	&	4.71	&	4.70	&	4.85	\\
& & MBPP & 3.08	&	4.11	&	4.11	&	4.19	&	4.07	&	4.34	\\
& & LiveCodeBench & 2.43	&	4.57	&	4.57	&	4.99	&	5.15	&	4.82	\\
\cmidrule{2-9}
& \multirow{4}{*}{Agentic} & SWE-Bench & 2.32	&	2.60	&	2.60	&	2.50	&	2.65	&	2.71	\\
& & Tau-Bench & 2.35	&	3.03	&	3.03	&	3.02	&	2.96	&	3.09	\\
& & AgentBench & 2.37	&	2.35	&	2.37	&	2.66	&	2.66	&	2.67	\\
& & AgentInstruct & 2.47	&	2.50	&	2.50	&	2.68	&	2.64	&	2.64	\\
\midrule
\multirow{15}{*}{Llama3.1-8B} & \multirow{4}{*}{Chat} & MT-Bench & 2.24	&	2.44	&	2.44	&	2.51	&	2.50	&	2.49	\\
& & Alpaca & 2.20	&	2.21	&	2.21	&	2.34	&	2.31	&	2.25	\\
& & UltraChat & 2.50	&	2.50	&	2.50	&	2.64	&	2.48	&	2.54	\\
& & WildChat & 1.97	&	2.05	&	2.05	&	2.18	&	2.08	&	2.18	\\
    \cmidrule{2-9}
& \multirow{4}{*}{Math} & GSM8K & 2.69	&	2.92	&	2.92	&	2.97	&	3.00	&	3.04	\\
& & MATH-500 & 2.45	&	2.92	&	2.92	&	2.99	&	2.79	&	2.73	\\
& & AIME24 & 2.25	&	2.82	&	2.82	&	2.89	&	2.84	&	2.70	\\
& & AIME25 & 2.29	&	2.86	&	2.86	&	2.93	&	2.93	&	3.06	\\
\cmidrule{2-9}
& \multirow{3}{*}{Coding} & HumanEval & 2.98	&	3.31	&	3.31	&	3.36	&	3.45	&	3.43	\\
& & MBPP & 2.90	&	3.16	&	3.16	&	3.22	&	3.33	&	3.29	\\
& & LiveCodeBench & 2.24	&	2.64	&	2.64	&	2.67	&	2.49	&	2.49	\\
\cmidrule{2-9}
& \multirow{4}{*}{Agentic} & SWE-Bench & 1.98	&	2.49	&	2.49	&	2.51	&	2.41	&	2.42	\\
& & Tau-Bench & 2.07	&	1.82	&	2.07	&	2.15	&	2.05	&	2.11	\\
& & AgentBench & 1.85	&	2.00	&	2.00	&	2.11	&	2.06	&	2.02	\\
& & AgentInstruct & 1.90	&	2.10	&	2.10	&	2.14	&	2.16	&	2.10	\\
    \bottomrule
  \end{tabular}
  \caption{Per-dataset speedup ($\times$) over vanilla AR decoding for three target models across 15 datasets from four task categories. Notably, the Oracle-Token speedup is omitted as it requires \texttt{float32} inference, preventing comparison with \texttt{bfloat16} inference.}
  \label{tab:app:speedup}
\end{table*}

\noindent
\section{Additional Analysis}\label{app:analysis}

In addition to the results presented in Section~\ref{sec:analysis}, we present the complete results for all the models employed in our study. 

\paragraph{Impact of Lazy Catch-up.}
Table~\ref{tab:app:lazy_catchup} shows the full throughput results regarding the impact of our proposed lazy catch-up spanning 15 datasets from four categories and three target models.

\begin{table*}[t]
  \small
  \centering
  \begin{tabular}{lll ccc}
    \toprule
    & & & \multicolumn{2}{c}{\textbf{Throughput}} &
    \\
    \cmidrule(lr){4-5}
    \textbf{Model} & \textbf{Category} & \textbf{Dataset} & \textbf{w/ lazy catch-up} & \textbf{w/o lazy catch-up} & \textbf{\% Increase} \\
    \midrule
\multirow{15}{*}{Qwen3-4B} & \multirow{4}{*}{Chat} & MT-Bench & 157.02	&	138.98	&	13.0\% \\
& & Alpaca & 140.55	&	119.44	&	17.7\% \\
& & UltraChat & 165.62	&	145.04	&	14.2\% \\
& & WildChat & 134.57	&	121.98	&	10.3\% \\
    \cmidrule{2-6}
& \multirow{4}{*}{Math} & GSM8K & 240.73	&	204.81	&	17.5\% \\
& & MATH-500 & 279.2	&	249.12	&	12.1\% \\
& & AIME24 & 265.37	&	247.91	&	7.0\% \\
& & AIME25 & 275.33	&	236.6	&	16.4\% \\
\cmidrule{2-6}
& \multirow{3}{*}{Coding} & HumanEval & 231.09	&	198.06	&	16.7\% \\
& & MBPP & 225.12	&	195.89	&	14.9\% \\
& & LiveCodeBench & 238.43	&	208.84	&	14.2\% \\
\cmidrule{2-6}
& \multirow{4}{*}{Agentic} & SWE-Bench & 138.78	&	124.59	&	11.4\% \\
& & Tau-Bench & 95.35	&	90.47	&	5.4\% \\
& & AgentBench & 139.68	&	120.09	&	16.3\% \\
& & AgentInstruct & 136.9	&	123.27	&	11.1\% \\
\midrule
\multirow{15}{*}{Qwen3-8B} & \multirow{4}{*}{Chat} & MT-Bench & 158.76	&	142.88	&	11.1\% \\
& & Alpaca & 144.3	&	133.7	&	7.9\% \\
& & UltraChat & 154.39	&	142.39	&	8.4\% \\
& & WildChat & 134.5	&	118.85	&	13.2\% \\
    \cmidrule{2-6}
& \multirow{4}{*}{Math} & GSM8K & 233.56	&	209.39	&	11.5\% \\
& & MATH-500 & 271.82	&	247.17	&	10.0\% \\
& & AIME24 & 276.59	&	244.41	&	13.2\% \\
& & AIME25 & 264.69	&	231.87	&	14.2\% \\
\cmidrule{2-6}
& \multirow{3}{*}{Coding} & HumanEval & 237.84	&	211.1	&	12.7\% \\
& & MBPP & 218.31	&	186.82	&	16.9\% \\
& & LiveCodeBench & 240.49	&	213.34	&	12.7\% \\
\cmidrule{2-6}
& \multirow{4}{*}{Agentic} & SWE-Bench & 131.82	&	120.66	&	9.2\% \\
& & Tau-Bench & 161	&	139.79	&	15.2\% \\
& & AgentBench & 129.13	&	114.39	&	12.9\% \\
& & AgentInstruct & 147.56	&	134.4	&	9.8\% \\
\midrule
\multirow{15}{*}{Llama3.1-8B} & \multirow{4}{*}{Chat} & MT-Bench & 177.39	&	163.39	&	8.6\% \\
& & Alpaca & 165.04	&	153.83	&	7.3\% \\
& & UltraChat & 179.06	&	166.05	&	7.8\% \\
& & WildChat & 156.39	&	130.21	&	20.1\% \\
    \cmidrule{2-6}
& \multirow{4}{*}{Math} & GSM8K & 199.98	&	176.63	&	13.2\% \\
& & MATH-500 & 191.18	&	166.56	&	14.8\% \\
& & AIME24 & 206.95	&	174.37	&	18.7\% \\
& & AIME25 & 211.76	&	185.68	&	14.0\% \\
\cmidrule{2-6}
& \multirow{3}{*}{Coding} & HumanEval & 222.58	&	201.05	&	10.7\% \\
& & MBPP & 234.23	&	200.36	&	16.9\% \\
& & LiveCodeBench & 177.28	&	155.93	&	13.7\% \\
\cmidrule{2-6}
& \multirow{4}{*}{Agentic} & SWE-Bench & 181.14	&	159.69	&	13.4\% \\
& & Tau-Bench & 152.76	&	135.2	&	13.0\% \\
& & AgentBench & 145.16	&	129.12	&	12.4\% \\
& & AgentInstruct & 148.93	&	125.02	&	19.1\% \\
    \bottomrule
  \end{tabular}
  \caption{Impact of lazy catch-up across all models and datasets. Deferring inactive-drafter KV updates to a single batched prefill at switch time improves throughput while leaving accepted tokens unchanged.}
  \label{tab:app:lazy_catchup}
\end{table*}

\paragraph{Switching Cost across All Models.} We present profiling results of switching cost (per cent w.r.t. per-round latency) with different prefill lengths and backlog token lengths for all the target models evaluated in our study. Table~\ref{tab:app:switching_cost} demonstrates that the cost stays below $7\%$ of per-round latency across all the models, and around 3--5\% at the representative 80--90 token backlog.

\begin{table}[t]
  \scriptsize
  \centering
  \begin{tabular}{ll ccccc}
    \toprule
  & & \multicolumn{5}{c}{\textbf{Backlog Token Length}} \\
  \cmidrule(lr){3-7}
  \textbf{Model}
  & \smash{\begin{tabular}[b]{@{}c@{}}\textbf{Prefill}\\\textbf{Tokens}\end{tabular}}
  & 1 & 64 & 128 & 512 & 1024 \\
    \midrule
    \multirow{6}{*}{Qwen3-4B} & 512 & 3.1\% & 3.6\% & 3.5\% & 3.5\% & 4.2\% \\
        & 1024 & 3.4\% & 3.5\% & 3.5\% & 3.5\% & 4.4\% \\
        & 2048 & 3.0\% & 3.5\% & 3.5\% & 3.8\% & 5.1\% \\
        & 4096 & 2.8\% & 3.7\% & 3.7\% & 4.1\% & 5.7\% \\
        & 8192 & 2.0\% & 3.4\% & 3.4\% & 3.9\% & 5.6\% \\
        & 16384 & 1.2\% & 3.0\% & 3.0\% & 3.5\% & 5.4\% \\
    \midrule
    \multirow{6}{*}{Qwen3-8B} & 512 & 3.3\% & 3.8\% & 3.8\% & 3.8\% & 4.9\% \\
        & 1024 & 3.3\% & 3.8\% & 3.8\% & 3.9\% & 5.2\% \\
        & 2048 & 3.2\% & 3.8\% & 3.8\% & 4.1\% & 5.6\% \\
        & 4096 & 2.8\% & 3.7\% & 3.7\% & 4.2\% & 5.8\% \\
        & 8192 & 1.9\% & 3.4\% & 3.4\% & 3.8\% & 5.7\% \\
        & 16384 & 1.2\% & 3.0\% & 3.0\% & 3.5\% & 5.4\% \\
    \midrule
    \multirow{6}{*}{Llama3.1-8B} & 512 & 4.3\% & 5.0\% & 4.9\% & 4.9\% & 6.5\% \\
        & 1024 & 4.2\% & 4.9\% & 4.8\% & 5.1\% & 6.8\% \\
        & 2048 & 3.8\% & 4.6\% & 4.6\% & 5.0\% & 6.8\% \\
        & 4096 & 3.2\% & 4.3\% & 4.3\% & 4.8\% & 6.7\% \\
        & 8192 & 2.2\% & 3.9\% & 3.8\% & 4.4\% & 6.4\% \\
        & 16384 & 1.4\% & 3.4\% & 3.4\% & 4.0\% & 6.1\% \\
    \bottomrule
  \end{tabular}
  \caption{Profiling results for switching cost (\% w.r.t. per-round latency) with different base KV lengths and backlog token lengths for all three target models (\textit{i.e.}~Qwen3-4B, Qwen3-8B, and Llama3.1-8B). The cost stays below 7\% of per-round latency across all the models, and around 3--5\% at the representative 80--90 token backlog.}
  \label{tab:app:switching_cost}
\end{table}

\paragraph{Memory Analysis across All Models.}
In addition to Table~\ref{tab:memory}, we further include peak memory measurements for Qwen3-4B and Llama3.1-8B as target models in Table~\ref{tab:app:memory}. This confirms our observation that the memory overhead of \ours is only marginally higher than that of DFlash, for different target models.

\paragraph{Robustness across Hardware.}
Table~\ref{tab:rtx4090_speedup} reports per-category speedup for all three target models on an NVIDIA RTX 4090, complementing the H100 results of Table~\ref{tab:throughput_main} and Section~\ref{sec:analysis}.

\begin{table}[th]
  \scriptsize
  \centering
  \addtolength{\tabcolsep}{-2pt}
  \begin{tabular}{ll cc}
    \toprule
    \textbf{Model} & \textbf{Method} & \textbf{Memory (GB)} & \textbf{\% Increase} \\
    \midrule
    \multirow{4}{*}{Qwen3-4B} & No SD & 7.55 & - \\
    & EAGLE-3           & 8.34 & 10.5\% \\
    & DFlash            & 8.96 & 18.8\% \\
    & \ours-Entropy/Neural & 9.38 & 24.3\% \\
    \midrule
    \multirow{4}{*}{Qwen3-8B} & No SD & 15.26 & \phantom{000}-- \\
    & EAGLE-3           & 16.39 & \phantom{0}7.4\% \\
    & DFlash            & 17.62 & 15.4\% \\
    & \ours-Entropy/Neural             & 18.41 & 20.6\% \\
    \midrule
    \multirow{4}{*}{Llama3.1-8B} & No SD & 14.96  & - \\
    & EAGLE-3           & 16.10 & 7.6\% \\
    & DFlash            & 17.27 & 15.5\% \\
    & \ours-Entropy/Neural & 18.06 & 20.7\% \\
    \bottomrule
  \end{tabular}
  \caption{Peak memory footprint of EAGLE-3, DFlash, and \ours with Qwen3-4B, Qwen3-8B, and Llama3.1-8B as target models. \ours's memory overheads over a diffusion-only system (DFlash) are $0.42$ GB, $0.79$ GB, and $0.79$ GB, merely $5.5\%$, $5.2\%$, and $5.2\%$ of the footprint of the target models, Qwen3-4B, Qwen3-8B and Llama3.1-8B, respectively.}
  \label{tab:app:memory}
\end{table}

\begin{table*}[t]
  \small
  \centering
  \begin{tabular}{ll ccccc}
    \toprule
    \textbf{Model} & \textbf{Method} & \textbf{Chat} & \textbf{Math} & \textbf{Coding} & \textbf{Agentic} & \textbf{Total Avg.} \\
    \midrule
    \multirow{6}{*}{Qwen3-4B} & EAGLE-3 & 2.41 & 2.65 & 2.58 & 1.83 & 2.35 \\
    & DFlash           & 1.96 & 4.55 & 4.10 & 2.05 & 3.10 \\
    \cmidrule{2-7}
    & Oracle-Task      & 2.41 & 4.55 & 4.10 & 2.05 & 3.22 \\
    & Oracle-Prompt    & 2.59 & 4.55 & 4.11 & 2.14 & 3.30 \\
    \cmidrule{2-7}
    & WhiFlash-Entropy & 2.48 & \textbf{4.52} & 4.09 & 2.11 & 3.25 \\
    & WhiFlash-Neural  & \textbf{2.56} & 4.49 & \textbf{4.11} & \textbf{2.16} & \textbf{3.28} \\
    \midrule
    \multirow{6}{*}{Qwen3-8B} & EAGLE-3 & 2.62 & 2.92 & 2.81 & 2.17 & 2.62 \\
    & DFlash           & 2.03 & 4.82 & 4.42 & 2.45 & 3.37 \\
    \cmidrule{2-7}
    & Oracle-Task      & 2.62 & 4.82 & 4.42 & 2.45 & 3.52 \\
    & Oracle-Prompt    & 2.80 & 4.83 & 4.43 & 2.58 & 3.61 \\
    \cmidrule{2-7}
    & WhiFlash-Entropy & 2.69 & 4.83 & 4.41 & 2.51 & 3.56 \\
    & WhiFlash-Neural  & \textbf{2.80} & \textbf{4.87} & \textbf{4.49} & \textbf{2.59} & \textbf{3.63} \\
    \midrule
    \multirow{6}{*}{Llama3.1-8B} & EAGLE-3 & 2.45 & 2.61 & 2.88 & 2.05 & 2.47 \\
    & DFlash           & 2.48 & 3.06 & 3.32 & 2.22 & 2.74 \\
    \cmidrule{2-7}
    & Oracle-Task      & 2.52 & 3.06 & 3.32 & 2.30 & 2.77 \\
    & Oracle-Prompt    & 2.63 & 3.13 & 3.35 & 2.35 & 2.83 \\
    \cmidrule{2-7}
    & WhiFlash-Entropy & 2.52 & \textbf{3.09} & \textbf{3.30} & 2.29 & \textbf{2.77} \\
    & WhiFlash-Neural  & \textbf{2.54} & 3.08 & 3.25 & \textbf{2.30} & 2.76 \\
    \bottomrule
  \end{tabular}
  \caption{Per-category speedup over vanilla AR decoding on an NVIDIA RTX 4090, for three target models across four task categories. Oracle rows are hindsight upper bounds. \textbf{Bold} denotes the best deployable method.}
  \label{tab:rtx4090_speedup}
\end{table*}

\end{document}